\documentclass{article}

\usepackage[nonatbib,final]{neurips_2020}
\usepackage[square,sort,comma,numbers]{natbib}
\usepackage{microtype}
\usepackage{xspace}
\usepackage{graphicx}
\usepackage{xcolor}
\usepackage{float}
\usepackage{subcaption}
\usepackage[T1]{fontenc}
\usepackage[utf8]{inputenc}
\usepackage{booktabs} 
\usepackage{hyperref}
\usepackage{amssymb}
\usepackage[tbtags]{amsmath}
\usepackage{hyperref}
\usepackage{textcomp}
\usepackage{gensymb}
\usepackage{algorithmic}
\usepackage{algorithm}

\usepackage[T1]{fontenc}
\usepackage[misc,geometry]{ifsym} 

\DeclareMathOperator*{\argmax}{argmax}

\newcommand{\indtable}[1]{\hspace{1.2em}#1}

\title{An artificial intelligence system for predicting the deterioration of COVID-19 patients in the emergency department}

\author{
Farah E. Shamout$^{1,*}$\textnormal{,}
Yiqiu Shen$^{2,*}$\textnormal{,}
Nan Wu$^{2,*}$\textnormal{,}
Aakash Kaku$^{2,*}$\textnormal{,}
Jungkyu Park$^{3,8*}$\textnormal{,}\\
\textbf{Taro Makino}$^{3,2,*}$,
\textbf{Stanisław Jastrzębski}$^{3,4,2}$,
\textbf{Jan Witowski}$^{3,4}$,
\textbf{Duo Wang}$^{5}$,
\textbf{Ben Zhang}$^{5}$,\\
\textbf{Siddhant Dogra}$^{3}$,
\textbf{Meng Cao}$^{6}$,
\textbf{Narges Razavian}$^{5,3,2}$,
\textbf{David Kudlowitz}$^{6}$,
\textbf{Lea Azour}$^{3}$,\\
\textbf{William Moore}$^{3}$,
\textbf{Yvonne W. Lui}$^{3,4}$,
\textbf{Yindalon Aphinyanaphongs}$^{5}$,\\
\textbf{Carlos Fernandez-Granda}$^{2,7}$,
\textbf{Krzysztof J. Geras}$^{3,4,2,\text{\Letter}}$\vspace{3mm}\\
$^1$Engineering Division, NYU Abu Dhabi\\
$^2$Center for Data Science, New York University\\
$^3$Department of Radiology, NYU Langone Health\\
$^4$Center for Advanced Imaging Innovation and Research, NYU Langone Health\\
$^5$Department of Population Health, NYU Langone Health\\
$^6$Department of Medicine, NYU Langone Health\\
$^7$Department of Mathematics, Courant Institute, New York University\\
$^8$Vilcek Institute of Graduate Biomedical Sciences, NYU Grossman School of Medicine\\
$^*$Equal contribution\\
$^\text{\Letter}$\texttt{k.j.geras@nyu.edu}
}

\begin{document}

\maketitle
\begin{abstract}
During the coronavirus disease 2019 (COVID-19) pandemic, rapid and accurate triage of patients at the emergency department is critical to inform decision-making. We propose a data-driven approach for automatic prediction of deterioration risk using a deep neural network that learns from chest X-ray images and a gradient boosting model that learns from routine clinical variables. Our AI prognosis system, trained using data from 3,661 patients, achieves an area under the receiver operating characteristic curve (AUC) of 0.786 (95\% CI: 0.745-0.830) when predicting deterioration within 96 hours. The deep neural network extracts informative areas of chest X-ray images to assist clinicians in interpreting the predictions and performs comparably to two radiologists in a reader study. In order to verify performance in a real clinical setting, we silently deployed a preliminary version of the deep neural network at New York University Langone Health during the first wave of the pandemic, which produced accurate predictions in real-time. In summary, our findings demonstrate the potential of the proposed system for assisting front-line physicians in the triage of COVID-19 patients.
\end{abstract}

\section{Introduction}
In recent months, there has been a surge in patients presenting to the emergency department (ED) with respiratory illnesses associated with the coronavirus disease 2019 (COVID-19)~\cite{baugh2020creating,debnath2020machine}. Evaluating the risk of deterioration of these patients to perform triage is crucial for clinical decision-making and resource allocation~\cite{whiteside2020redesigning}.
While ED triage is difficult under normal circumstances~\cite{dorsett2020point, mckenna2019emergency}, during a pandemic, strained hospital resources increase the challenge~\cite{warner2020stop,debnath2020machine}. This is compounded by our incomplete understanding of COVID-19. Data-driven risk evaluation based on artificial intelligence (AI) could, therefore, play an important role in streamlining ED triage. 

As the primary complication of COVID-19 is pulmonary disease, such as pneumonia~\cite{cozzi2020chest}, chest X-ray imaging is a first-line triage tool for COVID-19 patients~\cite{rubin2020role}. Although other imaging modalities, such as computed tomography (CT), provide higher resolution, chest X-ray imaging is less costly, inflicts a lower radiation dose, and is easier to obtain without incurring the risk of contaminating imaging equipment and disrupting radiologic services~\cite{american2020acr}. In addition, abnormalities in the chest X-ray images of COVID-19 patients have been found to mirror abnormalities in CT scans~\cite{wong2020frequency}. Although the knowledge of the disease is rapidly evolving, the understanding of the correlation between pulmonary parenchymal patterns visible in the chest X-ray images and clinical deterioration remains limited. This motivates the use of machine learning approaches for risk stratification using chest X-ray imaging, which may be able to learn such correlations automatically from data.

The majority of related previous work using imaging data of COVID-19 patients focus more on diagnosis than prognosis~\cite{kundu2020might,khan2020coronet,ucar2020covidiagnosis,li2020artificial, ozturk2020automated, wang2020fully, zhang2020clinically, singh2020classification}. Prognostic models used for predicting mortality, morbidity and other outcomes related to the disease course have a number of potential real-life applications, such as: consistently defining and triaging sick patients, alerting bed management teams on expected demands, providing situational awareness across teams of individual patients, and more general resource allocation~\cite{kundu2020might}. Prior methodology for prognosis of COVID-19 patients via machine learning mainly use routinely-collected clinical variables~\cite{wynants2020prediction, debnath2020machine} such as vital signs and laboratory tests, which have long been established as strong predictors of deterioration~\cite{news,shamout2019deep}. Some studies have proposed scoring systems for chest X-ray images to assess the severity and progression of lung involvement using deep learning~\cite{li2020automated}, or more commonly, through manual clinical evaluation~\cite{borghesi2020covid,toussie2020clinical,cozzi2020chest}. In general, the role of deep learning for the prognosis of COVID-19 patients using chest X-ray imaging has not yet been fully established. Using both the images and the clinical variables in a single AI system also has not been studied before. We show that they both contain complimentary information, which opens a new perspective on building prognostic AI systems for COVID-19.

In this work, we present an AI system that performs an automatic evaluation of deterioration risk, based on chest X-ray imaging, combined with other routinely collected non-imaging clinical variables. The goal is to provide support for critical clinical decision-making involving patients arriving at the ED in need of immediate care~\cite{debnath2020machine,fernandes2020clinical}, based on the need for efficient patient triage. The system is based on chest X-ray imaging, while also incorporating other routinely collected non-imaging clinical variables that are known to be strong predictors of deterioration.

Our AI system uses deep convolutional neural networks to perform risk evaluation from chest X-ray images. In particular, we designed our imaging-based classifier based on the Globally-Aware Multiple Instance Classifier (GMIC)~\cite{shen2019globally,shen2020interpretable}, denoted as COVID-GMIC, aiming for accurate performance and interpretability. The system also learns from routinely collected clinical variables using a gradient boosting model (GBM)~\cite{ke2017lightgbm}, denoted as COVID-GBM. Both models were trained using a dataset of 3,661 patients admitted to NYU Langone Health between March 3, 2020, and May 13, 2020. To learn from both modalities, we combined the output predictions of COVID-GMIC and COVID-GBM to predict each patient's overall risk of deterioration over different time horizons, ranging from 24 to 96 hours. In addition, the system includes a model that predicts how the risk of deterioration is expected to evolve over time by computing deterioration risk curves (DRC), in the spirit of survival analysis~\cite{miller2011survival}, denoted as COVID-GMIC-DRC.

Our system is able to accurately predict the deterioration risk on a test set of new patients. It achieves an area under the receiver operating characteristic curve (AUC) of 0.786 (95\% CI: 0.745-0.830), and an area under the precision-recall curve (PR AUC) of 0.517 (95\% CI: 0.429-0.600) for prediction of deterioration within 96 hours. Additionally, its estimated probability of the temporal risk evolution discriminates effectively between patients, and is well-calibrated. The imaging-based model achieves a comparable AUC to two experienced chest radiologists in a reader study, highlighting the potential of our data-driven approach. In order to verify our system's performance in a real clinical setting, we silently deployed a preliminary version of it at NYU Langone Health during the first wave of the pandemic, demonstrating that it can produce accurate predictions in real-time. Overall, these results strongly suggest that our system is a viable and valuable tool to inform triage of COVID-19 patients. For reproducibility, we published our code and the trained models at \href{https://github.com/nyukat/COVID-19_prognosis}{\url{https://github.com/nyukat/COVID-19_prognosis}}.

\begin{figure}[htb!]
    \centering
    \vspace{-4mm}
    \includegraphics[width=0.869\textwidth]{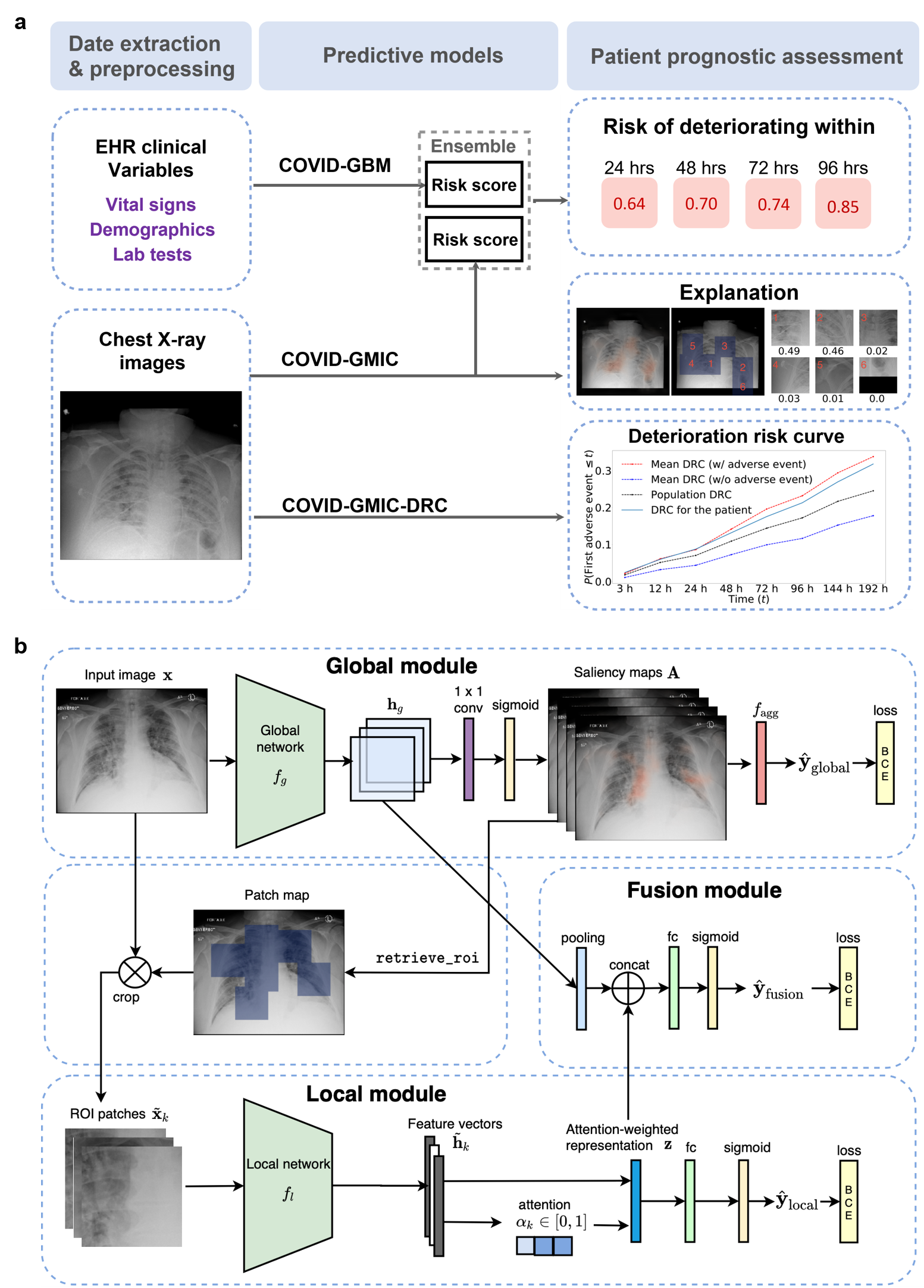}
    \caption{\small \textbf{Overview of the AI system and the architecture of its deep learning component.} \textbf{a}, Overview of the AI system that assesses the patient's risk of deterioration every time a chest X-ray image is collected in the ED. We design two different models to process the chest X-ray images, both based on the GMIC neural network architecture~\cite{shen2019globally,shen2020interpretable}. The first model, COVID-GMIC, predicts the overall risk of deterioration within 24, 48, 72, and 96 hours, and computes saliency maps that highlight the regions of the image that most informed its predictions. The predictions of COVID-GMIC are combined with predictions of a gradient boosting model~\cite{ke2017lightgbm} that learns from routinely collected clinical variables, referred to as COVID-GBM. The second model, COVID-GMIC-DRC, predicts how the patient's risk of deterioration evolves over time in the form of deterioration risk curves. \textbf{b}, Architecture of COVID-GMIC. First, COVID-GMIC utilizes the global network to generate four saliency maps that highlight the regions on the X-ray image that are predictive of the onset of adverse events within 24, 48, 72, and 96 hours respectively. COVID-GMIC then applies a local network to extract fine-grained visual details from these regions. Finally, it employs a fusion module that aggregates information from both the global context and local details to make a holistic diagnosis.}
    \label{fig:system-overview}
\end{figure}

\begin{figure}[ht]
    \centering
    \includegraphics[width=1\textwidth]{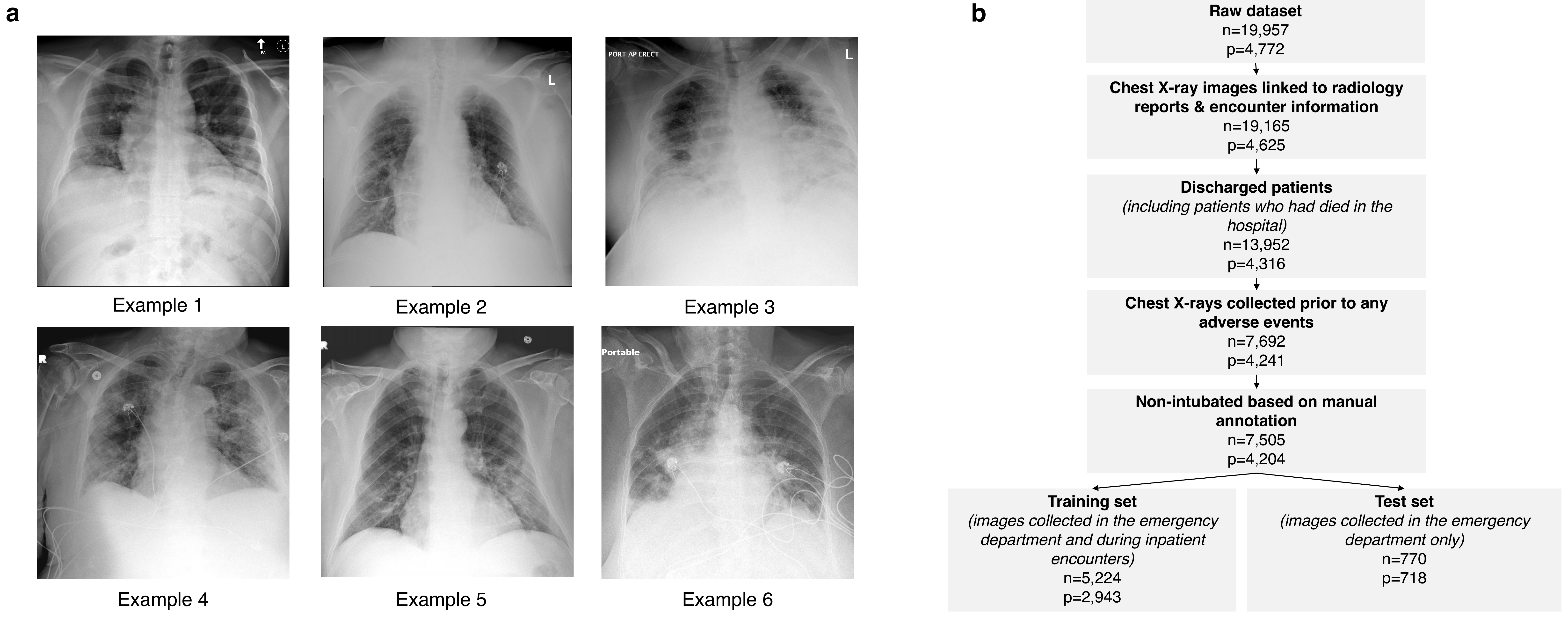}
    \caption{\small \textbf{Illustrations of the dataset and the dataset flowchart.} \textbf{a}, Examples of chest X-ray images in our dataset. Example 1: Patient was discharged and experienced no adverse events (44 years old male). Example 2: Patient was transferred to the ICU after 95 hours (71 years old male). Example 3: Patient was intubated after 72 hours (66 years old male). Example 4: Patient was transferred to the ICU after 48 hours (99 years old female). Example 5: Patient was intubated after 24 hours (74 years old male). Example 6: Patient was transferred to the ICU in 30 minutes (73 years old female). It is important to note that the extent of parenchymal disease does not necessarily have a direct correlation with deterioration time. For example, Example 5 has less severe parenchymal findings than Examples 3 and 4, but deteriorated faster. \textbf{b}, Flowchart showing how the inclusion and exclusion criteria were applied to obtain the final training and test sets, where $n$ represents the number of chest X-ray exams, and $p$ represents the number of unique patients. Specifically, we excluded 783 exams that were not linked to any radiology report, nine exams that had missing encounter information, and 5,213 exams from patients who were still hospitalised by May 13, 2020. To ensure that our system predicts deterioration prior to its occurrence, we excluded 6,260 exams that were collected after an adverse event and 187 exams of already intubated patients. The final dataset consisted of 7,502 chest X-ray exams corresponding to 4,204 unique patients. We split the dataset at the patient level such that exams from the same patient exclusively appear either in the training or test set. In the training set, we included exams that were collected both in the ED and during inpatient encounters. Since the intended clinical use of our model is in the ED, the test set only includes exams collected in the ED. This resulted in 5,224 exams (5,617 images) in the training set and 770 exams (832 images) in the test set. We included both frontal and lateral images, however there were less than 50 lateral images in the entire dataset. 
    }
    \label{fig:flowchart_examples}
\end{figure}

\section{Results}
\paragraph{Dataset.}
Our AI system was developed and evaluated using a dataset of 19,957 chest X-ray exams collected from 4,722 patients at NYU Langone Health between March 3, 2020 and May 13, 2020.\footnote{This study was approved by the Institutional Review Board, with ID\# i20-00858.} The dataset consists of chest X-ray images collected from patients who tested positive for COVID-19 using the polymerase chain reaction (PCR) test, along with the clinical variables recorded closest to the time of image acquisition (e.g. vital signs, laboratory test results, and patient characteristics). Figure~\ref{fig:flowchart_examples}.a shows examples of chest X-ray images collected from different patients. We applied inclusion and exclusion criteria that were defined in collaboration with clinical experts, as shown in Figure~\ref{fig:flowchart_examples}.b. The training set consisting of 2,943 patients and 5,617 chest X-ray images was used for model development and hyperparameter tuning, while the test set consisting of 718 patients and 832 images was used to report the final results. The training and the test sets were disjoint, with no patient overlap. Table~\ref{tab:data_stats} summarizes the overall demographics and characteristics of the patient cohort in the training and test sets, including distributions of the included clinical variables. The raw laboratory test variables were further processed to extract the minimum and maximum statistics. 

We define deterioration, the target to be predicted by our models, as the composite outcome of one of three adverse events: intubation, admission to the intensive care unit (ICU), and in-hospital mortality. If multiple adverse events occurred, we only consider the time of the first event. 

\begin{table}[ht!]
\centering
\small
\caption{\small Description of the characteristics of the patient cohort included in the training and test sets and the mean and interquartile range statistics of the raw vital signs and laboratory test results used for COVID-GBM. Note that $n$ represents a counting unit. The training and test sets are similar in terms of age, BMI, and proportion of females. We note that there is a higher proportion of chest X-ray images associated with deterioration across all time windows in the test set compared to the training set. This implies that there is a higher incidence of adverse events amongst ED patients than inpatients, since the test set only includes chest X-ray images collected from ED patients, while the training set also includes inpatients.}
\begin{tabular}{@{}lll@{}}  
    \toprule 
    Characteristic & Training set & Test set \\
   \midrule 
   
   Patients, n & 2,943 & 718 \\
   \indtable{Admissions, n} & 3,175 & 764 \\
   \indtable{Females, n (\%)} & 1,206 (41.0) & 305 (42.5) \\
   \indtable{Age (years)}, mean (SD) & 62.9 (17.2)&  64.9 (17.2) \\
   \indtable{BMI (kg/m$^2$)}, mean (SD) & 29.4 (7.0) & 29.5 (8.6) \\
   \indtable{Survived} & 2,405 & 559 \\ 
   
   Adverse events, n & 1,311 & 594 \\
   \indtable{Intubation, n} & 386 & 97 \\
   \indtable{ICU admission, n} & 387 & 113 \\
   \indtable{Mortality, n} & 538 & 159 \\
   \indtable{Composite outcome, n} & 730 & 225 \\ 
   
   Chest X-ray exams, n & 5,224 & 770\\
   \indtable{Composite outcome within 24 hours, n (\%)} & 349 (6.7\%) & 74 (9.6\%)\\
   \indtable{Composite outcome within 48 hours, n (\%)} & 553 (10.6\%) & 101 (13.1\%)\\
   \indtable{Composite outcome within 72 hours, n (\%)} & 735 (14.1\%) & 130 (16.9\%) \\
   \indtable{Composite outcome within 96 hours, n (\%)} & 876 (16.8\%) & 156 (20.3\%)\\
   \indtable{Total number of images, n} & 5,617 & 832\\
   
   Vital signs, \textit{units} 
   \\
   \indtable{Heart rate, \textit{beats per minute}} & 93.7 (25.0) & 93.5 (27.0)\\
      \indtable{Respiratory rate, \textit{breaths per minute}} & 22.4 (7.0)& 23.4 (7.0)\\
   \indtable{Temperature, \textit{$\degree$F}} & 99.4 (1.9) & 99.4 (1.9)\\
   \indtable{Systolic blood pressure, \textit{mmHg}} & 130.7 (30.0) & 129.8 (29.3) \\
   \indtable{Diastolic blood pressure,  \textit{mmHg}} & 75.9 (17.0) & 76.0 (18.0)\\
   \indtable{Oxygen saturation,  \textit{\%}} & 94.1 (4.0) & 93.8 (5.0) \\

    Laboratory tests, \textit{units} \\
     \indtable{Albumin, \textit{g/dL}}& 3.5 (0.9) & 3.5 (0.9)\\
     \indtable{ALT, \textit{U/L}} & 49.8 (32.0) & 52.2 (36.0)\\
     \indtable{AST, \textit{U/L}} & 67.3 (37.0) & 69.7 (43.0)\\
     \indtable{Total bilirubin, \textit{mg/dL}} & 0.7 (0.4) &  0.7 (0.4)\\
     \indtable{Blood urea nitrogen, \textit{mg/dL}} & 25.9 (17.0) & 26.4 (18.0) \\
     \indtable{Calcium, \textit{mg/dL}} & 8.7 (0.8) & 8.7 (0.8) \\
     \indtable{Chloride, \textit{mEq/L}} & 101.1 (7.0) & 101.6 (7.0)\\
     \indtable{Creatinine, \textit{mg/dL}} & 1.6 (0.7) &  1.6 (0.7) \\
     \indtable{D-dimer, \textit{ng/mL}} & 1,321.6 (535.5) & 1,146.3 (618.5) \\
    \indtable{Eosinophils, \textit{\%}} & 0.4 (0.0) & 0.4 (0.0) \\
     \indtable{Eosinophils, \textit{n}} & 0.03 (0.00) & 0.03 (0.00) \\
    \indtable{Hematocrit, \textit{\%}} & 38.9 (7.3) & 38.9 (7.5) \\
     \indtable{LDH, \textit{U/L}} & 412.8 (207.0) & 404.0 (213.0) \\
     \indtable{Lymphocytes, \textit{\%}} & 14.1 (10.0) & 14.9 (11.0) \\
     \indtable{Lymphocytes, \textit{n}} & 1.0 (0.7) & 1.0 (0.7)\\
     \indtable{Platelet volume, \textit{fL}} & 10.6 (1.4) & 10.6 (1.4) \\
     \indtable{Neutrophils, \textit{n}} & 6.4 (4.0) & 6.3 (3.8)\\
     \indtable{Neutrophils, \textit{\%}} & 76.6 (14.0) & 75.9 (13.0)  \\
     \indtable{Platelet, \textit{n}} & 226.1 (114.0) & 223.7 (103.0) \\
     \indtable{Potassium, \textit{mmol/L}} & 4.2 (0.8) & 4.2 (0.8)\\
     \indtable{Procalcitonin, \textit{ng/mL}} & 1.9 (0.3) & 1.9 (0.4)  \\
     \indtable{Total protein, \textit{g/dL}} & 7.1 (1.1) & 7.2 (1.0)\\
     \indtable{Sodium, \textit{mmol/L}} & 136.2 (6.0) & 136.6 (7.0) \\
     \indtable{Troponin, \textit{ng/mL}} & 0.2 (0.1) & 0.2 (0.1)\\
    \bottomrule
    \end{tabular}
\label{tab:data_stats}
\end{table}

\paragraph{Evaluation metrics.} Throughout the paper we used AUC (area under the receiver operating characteristic curve) and PR AUC (area under the precision-recall curve), which offer a complimentary view on the performance of our models. These metrics integrate the performance of the evaluated models over all possible thresholds for predictions to be considered positive. As there are no available guidelines on how to select the threshold, we prefer these metrics to metrics that are computed for a fixed threshold (i.e.,  F1 score or classification accuracy).

\paragraph{Model performance.}
Table~\ref{tab:overall-performance} summarizes the performance of all the models in terms of the AUC and PR AUC for the prediction of deterioration within 24, 48, 72, and 96 hours from the time of the chest X-ray exam. The receiver operating characteristic curves and precision-recall curves can be found in Supplementary Figure~\ref{fig:all_peformance_task_1}. For comparison, a logistic regression model achieves 0.698,  0.699, 0.712, and 0.728 AUC and 0.214, 0.266, 0.339, and 0.436 PR AUC across the 24, 48, 72, and 96 hours windows, respectively. Our ensemble model consisting of COVID-GMIC and COVID-GBM achieves the best AUC performance across all time windows compared to COVID-GMIC and COVID-GBM individually. This highlights the complementary role of chest X-ray images and routine clinical variables in predicting deterioration. The weighting of the predictions of COVID-GMIC and COVID-GBM was optimized on the validation set, as shown in Supplementary Figure~\ref{fig:top-features-xgboost-validation-ensemble}.b. Similarly, the ensemble of COVID-GMIC and COVID-GBM outperforms all models across all time windows in terms of the PR AUC, except for the 96 hours window.
The consistent advantage of the ensemble model in our results is especially encouraging. Investigating more complex strategies for fusion of information from these two modalities could further improve the results and this will be a subject of our future research. Sample learning curves of COVID-GMIC are shown in Supplementary Figure~\ref{fig:learning_curves} for reference.

\begin{table}[ht]
\centering
\caption{\small Performance of the outcome classification task on the held-out test set, and on the subset of the test set used in the reader study. We include 95\% confidence intervals estimated by 1,000 iterations of the bootstrap method~\cite{efron1994introduction}. The optimal weights assigned to the COVID-GMIC prediction in the COVID-GMIC and COVID-GBM ensemble were derived through optimizing the AUC on the validation set as described in Supplementary Figure~\ref{fig:top-features-xgboost-validation-ensemble}.b. The ensemble of COVID-GMIC and COVID-GBM, denoted as `COVID-GMIC + COVID-GBM', achieves the best performance across all time windows in terms of the AUC and PRAUC, except for the PR AUC in the 96 hours task. In the reader study, our main finding is that COVID-GMIC outperforms radiologists A \& B across time windows longer than 24 hours, with 3 and 17 years of experience, respectively. Note that the radiologists did not have access to clinical variables and as such their performance is not directly comparable to the COVID-GBM model; we include it only for reference. The area under the precision-recall curve is sensitive to class distribution, which explains the large differences between the scores on the test set and the reader study subset.}
\resizebox{.995\textwidth}{!}{\begin{tabular}{@{}lccccccccc@{}}  
    \toprule 
    \multicolumn{10}{c}{\textbf{\large Test set (n=832)}}\\
    \midrule
    & \multicolumn{4}{c}{AUC} & \phantom{a} & \multicolumn{4}{c}{PR AUC}\\
    \cmidrule{2-5} \cmidrule{7-10}
      &24 hours &48 hours &72 hours &96 hours  & \phantom{a}&24 hours &48 hours &72 hours &96 hours\\
  \midrule 
  COVID-GBM & 0.747  & 0.739  & 0.750  & 0.770  & \phantom{a} & 0.230  & 0.325  & 0.408  & \textbf{0.523} \\
  & (0.698, 0.802) & (0.69, 0.795) & (0.703, 0.799) & (0.727, 0.813) &  \phantom{a}  & (0.139, 0.296) & (0.229, 0.396) & (0.317, 0.479) & (0.433, 0.6)\\ \midrule
COVID-GMIC    &   0.695  & 0.716  & 0.717  & 0.738  & \phantom{a}&  0.200  & 0.302   & 0.374    & 0.439   \\
 & (0.636, 0.763) & (0.666, 0.771) & (0.668, 0.773) & (0.695, 0.785) &  \phantom{a} & (0.119, 0.260) & (0.209, 0.379) & (0.283, 0.452) & (0.346, 0.515)\\ \midrule
COVID-GBM + &  \textbf{0.765}   & \textbf{0.749}  & \textbf{0.769} &  \textbf{0.786}  & \phantom{a} & \textbf{0.243}  & \textbf{0.332}   & \textbf{0.439}   & 0.517  \\
COVID-GMIC  & (0.712, 0.817) & (0.700, 0.798)  & (0.724, 0.818) & (0.745, 0.830) &  \phantom{a}  & (0.150, 0.299) & (0.237, 0.41) & (0.345, 0.527) & (0.429, 0.600) \\
\bottomrule
\toprule 
   \multicolumn{10}{c}{\textbf{\large Reader study dataset (n=200)}}\\
  \midrule
 &\multicolumn{4}{c}{AUC} & \phantom{a} & \multicolumn{4}{c}{PR AUC}\\
    \cmidrule{2-5} \cmidrule{7-10}
     &24 hours &48 hours &72 hours &96 hours  & \phantom{a}&24 hours &48 hours &72 hours &96 hours\\
  \midrule 
Radiologist A &  0.613  &  0.645  &  0.691  &  0.740  & \phantom{a}&    0.346  &  0.490  &  0.640  &  0.742 \\ 
& (0.519, 0.705) & (0.571, 0.731) & (0.618, 0.77) & (0.674, 0.814) &  \phantom{a} & (0.217, 0.441) & (0.367, 0.599) & (0.536, 0.745) & (0.657, 0.834)\\ \midrule
Radiologist B     &   0.637  &  0.636  &  0.658  &  0.713 & \phantom{a}&    0.365  &  0.460  &  0.590  &  0.704 \\ 
  & (0.547, 0.73) & (0.552, 0.716) & (0.588, 0.738) & (0.649, 0.786) &  \phantom{a} & (0.229, 0.462) & (0.335, 0.56) & (0.492, 0.701) & (0.616, 0.805)\\ \midrule
Radiologist A + &  \textbf{0.642} &  0.663  &  0.692  &  0.741& \phantom{a}&    \textbf{0.403}  &  0.499  &  0.609  &  0.740 \\ 
Radiologist B  &   (0.555, 0.729)  & (0.589, 0.746) & (0.621, 0.766) & (0.678, 0.809) &  \phantom{a} & (0.272, 0.52) & (0.380, 0.613) & (0.492, 0.711) & (0.650, 0.831)\\ \midrule
COVID-GMIC   &  \textbf{0.642}  &  \textbf{0.701}  &  \textbf{0.751} &  \textbf{0.808} & \phantom{a} &   0.381  &  \textbf{0.546} &  \textbf{0.676}  &  \textbf{0.789} \\
  & (0.554, 0.734) & (0.627, 0.781) & (0.685, 0.821) & (0.75, 0.87) &  \phantom{a}  & (0.235, 0.480) & (0.421, 0.657) & (0.564, 0.780) & (0.699, 0.880)\\
  \toprule 
COVID-GBM    &  0.704  &  0.719  &  0.750  &  0.787 & \phantom{a} &    0.411  &  0.537  &  0.668  &  0.804 \\
  & (0.632, 0.784) & (0.648, 0.794) & (0.684, 0.821) & (0.727, 0.850) &  \phantom{a}  & (0.259, 0.518) & (0.394, 0.64) & (0.558, 0.77) & (0.738, 0.884)\\ \midrule
COVID-GBM + &  0.708  &  0.702  &  0.778  &  0.819 & \phantom{a} &  0.411 &  0.500  &  0.705  &  0.808\\
COVID-GMIC  & (0.637, 0.799) & (0.633, 0.775) & (0.719, 0.851) & (0.763, 0.885) &  \phantom{a}  & (0.279, 0.517) & (0.364, 0.601) & (0.599, 0.806) & (0.735, 0.898)\\
    \bottomrule \\
\end{tabular}}
\label{tab:overall-performance}
\end{table}

\begin{figure}[ht]
    \centering
    \includegraphics[width=0.99\textwidth]{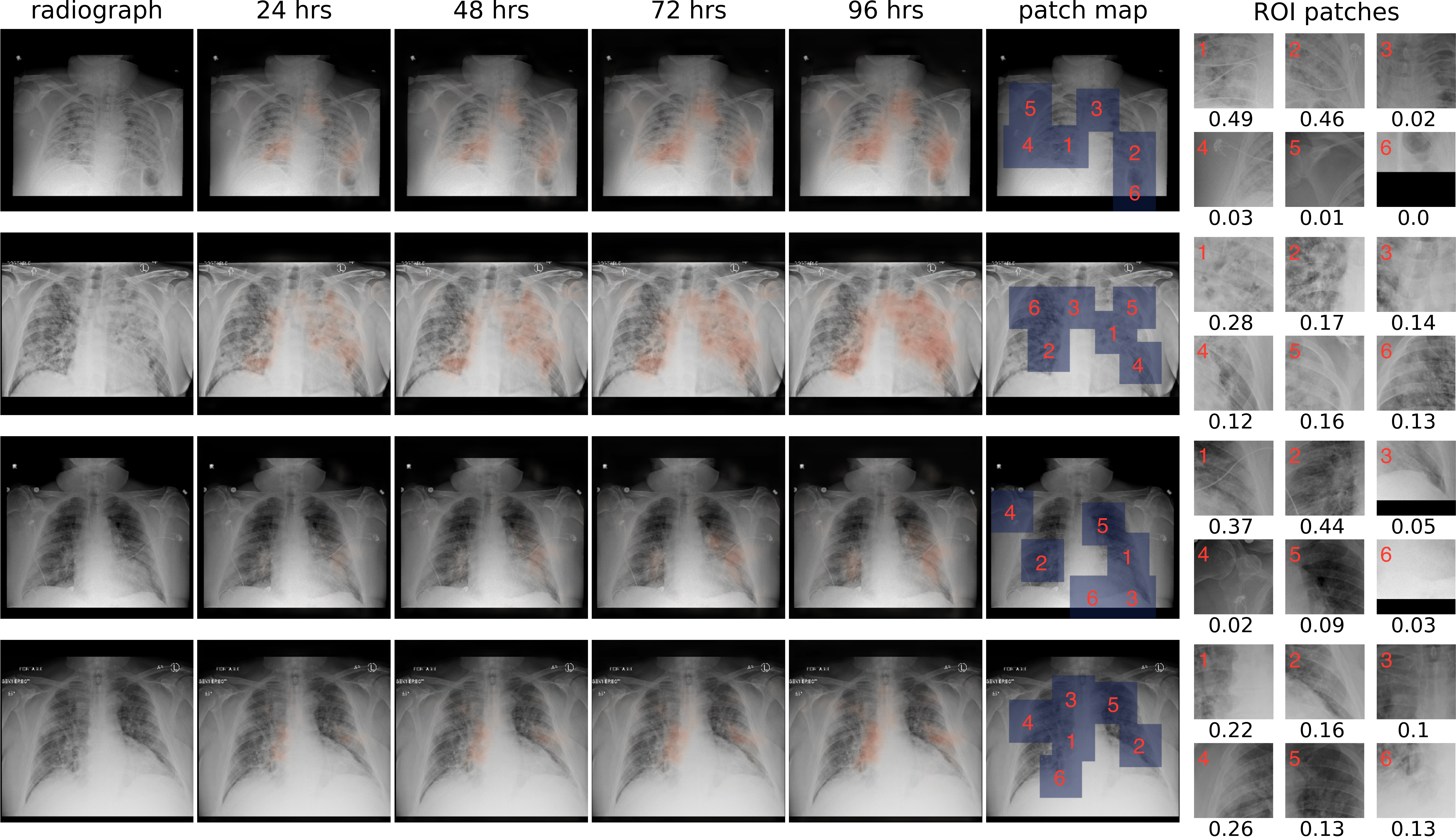}\\ \vspace{-5pt}
    \caption{\small \textbf{Explainability of COVID-GMIC.} From left to right: the original X-ray image, saliency maps for clinical deterioration within 24, 48, 72, and 96 hours, locations of region-of-interest (ROI) patches, and ROI patches with their associated attention scores. All four patients were admitted to the intensive care unit and were intubated within 48 hours. In the first example, there are diffuse airspace opacities, though the saliency maps primarily highlight the medial right basilar and peripheral left basilar opacities. Similarly, the two ROI patches (1 and 2) on the basilar region demonstrate comparable attention values, 0.49 and 0.46 respectively. In the second example, the extensive left mid to upper-lung abnormality (ROI patch 1) is highlighted, which correlates with the most extensive area of parenchymal consolidation. In the third example, the saliency maps highlight the left mid lung (ROI patch 1) and right hilar/infrahilar regions (ROI patch 2) which show groundglass opacities. In the last example, saliency maps highlight the right mid to lower paramediastinal (ROI patch 1 and 6) and left mid lung periphery (ROI patch 2) as regions predictive of clinical deterioration within 96 hours.}
    \label{fig:gmic_vis}
\end{figure}

To illustrate the interpretability of COVID-GMIC, we show in Figure~\ref{fig:gmic_vis} the saliency maps for all time windows (24, 48, 72, and 96 hours) computed for four examples from the test set. Across all four examples, the saliency maps highlight regions that contain visual patterns such as airspace opacities and consolidation, which are correlated with clinical deterioration~\cite{li2020automated,toussie2020clinical}. These saliency maps are utilized to guide the extraction of six regions of interest (ROI) patches cropped from the entire image, which are then associated with a score that indicates its relevance to the prediction task. We also note that in the last example, the saliency maps highlight right mid to lower paramediastinal and left mid-lung periphery, while missing the dense consolidation in the periphery of the right upper lobe. This suggests that COVID-GMIC emphasizes only the most informative regions in the image, while human experts can provide a more holistic interpretation covering the entire image. It might, therefore, be useful to enhance GMIC through a classifier agnostic mechanism~\cite{zolna2020classifier}, which finds all the useful evidence in the image, instead of solely the most discriminative part. We leave this for future work.

\paragraph{Comparison to radiologists.}
We compared the performance of COVID-GMIC with two chest radiologists from NYU Langone Health (with 3 and 17 years of experience) in a reader study with a sample of 200 frontal chest X-ray exams from the test set.  We used stratified sampling to improve the representation of patients with a negative outcome in the reader study dataset.  Specifically, we randomly sampled the first 100 exams from patients that had an adverse event in the next 96 hours from the time the exam was taken. The remaining 100 exams came from the complement of the test set. We describe the design of the reader study in more detail in the Methods section. 

As shown in Table~\ref{tab:overall-performance}, our main finding is that COVID-GMIC achieves a comparable performance to radiologists across all time windows in terms of AUC and PR AUC, and outperforms radiologists for 48, 72, and 96 hours. For example, COVID-GMIC achieves AUC of 0.808 (95\% CI, 0.746-0.866) compared to AUC of 0.741 average AUC of both radiologists in the 96 hours prediction task. We hypothesize that COVID-GMIC outperforms radiologists on this task due to the currently limited clinical understanding of which pulmonary parenchymal patterns predict clinical deterioration, rather than the severity of lung involvement~\cite{toussie2020clinical}. Supplementary Figure~\ref{fig:reader_study_roc_pr} shows AUC and PR AUC curves across all time windows. 

\paragraph{Deterioration risk curves.} \label{sec:results_4_risk_profile}

\begin{figure}[h!]
    \includegraphics[width=\textwidth]{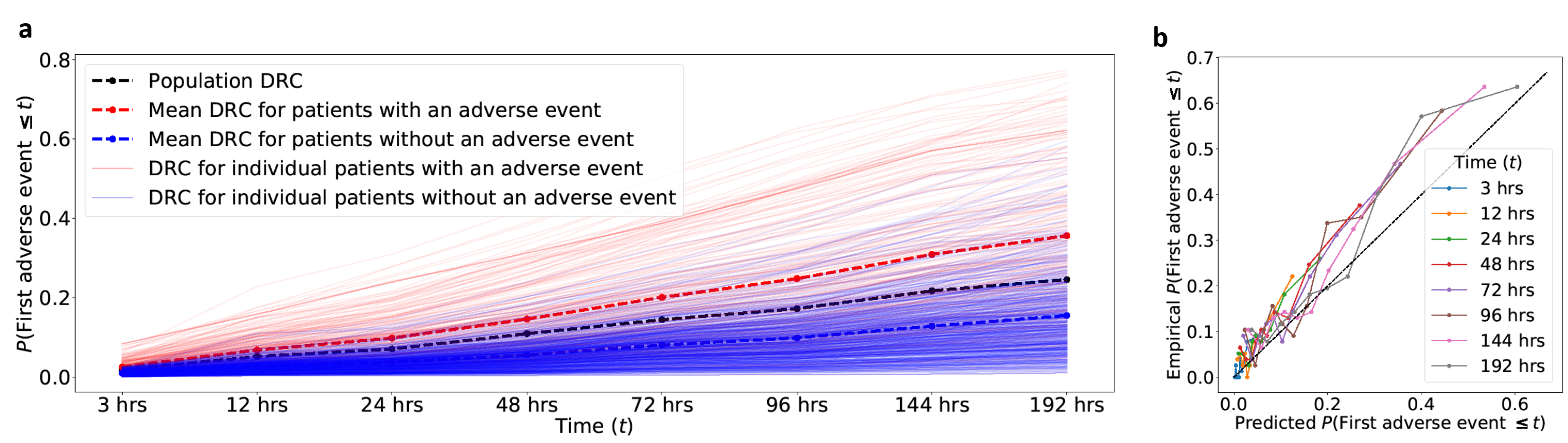}
    \caption{\small\textbf{Deterioration risk curves (DRCs) and reliability plot for COVID-GMIC-DRC.} \textbf{a}, DRCs generated by the COVID-GMIC-DRC model for patients in the test set with (faded red lines) and without adverse events (faded blue lines). The mean DRC for patients with adverse events (red dashed line) is higher than the DRC for patients without adverse events (blue dashed line) at all times. The graph also includes the ground-truth population DRC (black dashed line) computed from the test data. 
    \textbf{b}, Reliability plot of the DRCs generated by the COVID-GMIC-DRC model for patients in the test set. The empirical probabilities are computed by dividing the patients into deciles according to the value of the DRC at each time $t$. The empirical probability equals the fraction of patients in each decile that suffered adverse events up to $t$. This is plotted against the predicted probability, which equals the mean DRC of the patients in the decile. Perfect calibration is indicated by the diagonal black dashed line.
    }
    \label{fig:gmic_survival_model}
\end{figure}

We use a modified version of COVID-GMIC, referred to hereafter as COVID-GMIC-DRC, to generate discretized deterioration risk curves (DRCs) which predict the evaluation of the deterioration risk based on chest X-ray images. Figure~\ref{fig:gmic_survival_model}.a shows the DRCs for all the patients in the test set. The DRC represents the probability that the first adverse event occurs before time $t$, where $t$ is equal to 3, 12, 24, 48, 72, 96, 144, and 192 hours. The mean DRCs of patients who deteriorate (red bold line) is significantly higher than the mean DRCs of patients who are discharged without experiencing any adverse events (blue bold line). We evaluate the performance of the model using the concordance index, which is computed on patients in the test set who experienced adverse events. For a fixed time $t$ the index equals the fraction of pairs of patients in the test data for which the patient with the higher DRC value at $t$ experiences an adverse event earlier. For $t$ equal to 96 hours, the concordance index is 0.713 (95\% CI: 0.682-0.747), which demonstrates that COVID-GMIC-DRC can effectively discriminate between patients. Other values of $t$ yield similar results, as reported in Supplementary Table~\ref{tab:survival_model_appendix}. Sample learning curves of COVID-GMIC-DRC are shown in Supplementary Figure~\ref{fig:learning_curves_drc} for reference.

Figure~\ref{fig:gmic_survival_model}.b shows a reliability plot, which evaluates the calibration of the probabilities encoded in the DRCs. The diagram compares the values of the estimated DRCs for the patients in the test set with empirical probabilities that represent the true frequency of adverse events. To compute the empirical probabilities, we divided the patients into deciles according to the value of the DRC at each time $t$. We then computed the fraction of patients in each decile that suffered adverse events up to $t$. The fraction is plotted against the mean DRC of the patients in the decile. The diagram shows that these values are similar across the different values of $t$, meaning the model is well-calibrated (for comparison, perfect calibration would correspond to the diagonal black dashed line).

\paragraph{Prospective silent validation in a clinical setting.}
Our long-term goal is to deploy our system in existing clinical workflows to assist clinicians. The clinical implementation of machine learning models is a very challenging process, both from technical and organizational standpoints~\citep{baier2019challenges}. To test the feasibility of deploying the AI system in the hospital, we silently deployed a preliminary version of our AI system in the hospital system and let it operate in real-time beginning on May 22, 2020. The deployed version includes 15 models that are based on DenseNet-121 architectures, and use only chest X-ray images. The models were developed to predict deterioration within 96 hours using a subset of our data collected prior to deployment from 3,425 patients. The models were serialized and served with TensorFlow Serving components~\cite{tensorflow2015-whitepaper} on an Intel(R) Xeon(R) Gold 6154 CPU @ 3.00GHz; no GPUs were used. Images are preprocessed as explained in the Methods section. Our system produces predictions essentially in real-time - it takes approximately two seconds to extract an image from the PACS system, apply the image preprocessing steps, and get the prediction of a model as a TensorFlow~\cite{tensorflow2015-whitepaper} output.

A total of 375 exams were collected between May 22, 2020 and June 24, 2020. Of the 375 exams collected between May 22, 2020 and June 24, 2020, 38 exams were associated with a positive 96 hour deterioration outcome. An ensemble of the deployed models, obtained by averaging their predictions, achieved an AUC of 0.717 (95\% CI: 0.622-0.801) and a PR AUC of 0.289 (95\% CI: 0.181-0.465). These results are comparable to those obtained on a retrospective test set used for evaluation before deployment, which are 0.748 (95\% CI: 0.708-0.790) AUC and 0.365 (95\% CI: 0.313-0.465) PR AUC. The decrease in accuracy is expected and may indicate changes in the patient population and treatment guidelines as the pandemic progressed. When practically deployed, our system would still need periodical retraining with the latest data.

\section{Discussion}
In this work, we present an AI system that is able to predict deterioration of COVID-19 patients presenting to the ED, where deterioration is defined as the composite outcome of mortality, intubation, or ICU admission. The system aims to provide clinicians with a quantitative estimate of the risk of deterioration, and how it is expected to evolve over time, in order to enable efficient triage and prioritization of patients at the high risk of deterioration. The tool may be of particular interest for pandemic hotspots where triage at admission is critical to allocate limited resources such as hospital beds.

Recent studies have shown that chest X-ray images are useful for the diagnosis of COVID-19~\cite{wynants2020prediction,narin2020automatic,khan2020coronet,ucar2020covidiagnosis,ozturk2020automated}. Our work supplements those studies by demonstrating the significance of this imaging modality for COVID-19 prognosis. Additionally, our results suggest that chest X-ray images and routinely collected clinical variables contain complementary information, and that it is best to use both to predict clinical deterioration. This builds upon existing prognostic research, which typically focuses on developing risk prediction models using non-imaging variables extracted from electronic health records~\cite{shamout2020review,wynants2020prediction}. In Supplementary Table~\ref{tab:trainingsize}, we demonstrate that our models' performance can be improved by increasing the dataset size. The current dearth of prognosis models that use both imaging and clinical variables may partly be due to the limited availability of large-scale datasets including both data types and outcome labels, which is a key strength of our study. In order to assess the clinical benefits of our approach, we conducted a reader study, and the results indicate that the proposed system can perform comparably to radiologists. This highlights the potential of data-driven tools for assisting the interpretation of X-ray images.

The proposed deep learning model, COVID-GMIC, provides visually intuitive saliency maps to help clinicians interpret the model predictions~\cite{ahmad2018interpretable}.Existing work on COVID-19 often use external gradient-based algorithms, such as gradCAM~\cite{selvaraju2017grad}, to interpret deep neural network classifiers~\cite{song2019current,brunese2020explainable,paul2020generalizability}. However, visualizations generated by gradient-based methods are sensitive to minor perturbation in input images, and could yield misleading interpretations~\cite{adebayo2018sanity}. In contrast, COVID-GMIC has an inherently interpretable architecture that better retains localization information of the more informative regions in the input images.

We performed prospective validation of an early version of our system through silent deployment in an NYU Langone Health hospital. The results suggest that the implementation of our AI system in the existing clinical workflows is feasible. Our model does not incur any overhead operational costs on data collection, since chest X-ray images are routinely collected from COVID-19 patients. Additionally, the model can process the image efficiently in real-time, without requiring extensive computational resources such as GPUs. This is an important strength of our study, since very few studies have implemented and prospectively validated risk prediction models in general~\cite{brajer2020prospective}. 

Our approach has some limitations that will be addressed in future work. The silent deployment was based only on the model that processes chest X-ray exams, and did not include routine clinical variables, nor any interventions. In addition, further validation is required to assess whether the system can improve key performance measures, such as patient outcomes, through prospective and external validation across different hospitals and electronic health records systems.

Our system currently considers two data types, which are chest X-ray images and clinical variables. Incorporating additional data from patient health records may further improve its performance. For example, the inclusion of presenting symptoms using natural language processing has been shown to improve the performance of a risk prediction model in the ED~\cite{fernandes2020clinical}. Although we focus on chest X-ray images because pulmonary disease is the main complication associated with COVID-19, COVID-19 patients may also suffer poor outcomes due to non-pulmonary complications such as: non-pulmonary thromboembolic events, stroke, and pediatric inflammatory syndromes~\cite{lodigiani2020venous,oxley2020large,viner2020kawasaki}. This could explain some of the false negatives incurred by our system; therefore, incorporating other types of data that reflect non-pulmonary complications may also improve prognostic accuracy.

Our system was developed and evaluated using data collected from the NYU Langone Health in New York, USA. Therefore, it is possible that our models overfit to the patient demographics and specific configurations in the imaging acquisition devices of our dataset. 

Our findings show the promise of data-driven AI systems in predicting the risk of deterioration for COVID-19 patients, and highlights the importance of designing multi-modal AI systems capable of processing different types of data. We anticipate that such tools will play an increasingly important role in supporting clinical decision-making in the future. 

\section{Methods}
\paragraph{Outline.}
In this section, we first introduce our image preprocessing pipeline then formulate the adverse event prediction task and present our multi-modal approach which utilizes both chest X-ray images and clinical variables. Next, we formally define deterioration risk curve (DRC) and introduce our X-ray image-based approach to estimate DRC. Subsequently, we summarize the technical details of model training and implementation. Lastly, we describe the design of the reader study.

\paragraph{Image Preprocessing.} 

After extracting the images from DICOM files, we applied the following preprocessing procedure. We first thresholded and normalized pixel values, and then cropped the images to remove any zero-valued pixels surrounding the image. Then, we unified the dimensions of all images by cropping the images outside the center and rescaling. We performed data augmentation by applying random horizontal flipping ($p=0.5$), random rotation (-45 to 45 degrees), and random translation. Supplementary Figure~\ref{fig:preprocessing} shows the distribution of the size of the images prior to data augmentation, as well as examples of images before and after preprocessing.

\paragraph{Adverse event prediction.}
Our main goal is to predict clinical deterioration within four time windows of 24, 48, 72, and 96 hours. We frame this as a multi-label classification task with binary labels $\mathbf{y} = [y^{24},y^{48}, y^{72}, y^{96}]$ indicating clinical deterioration of a patient within the four time windows. The probability of deterioration is estimated using two types of data associated with the patient: a chest X-ray image, and routine clinical variables. We use two different machine learning models for this task: COVID-GMIC to process chest X-ray images, and COVID-GBM to process clinical variables. For each time window $t \in \mathbb{T}_a = \{24, 48, 72, 96\}$, both models produce probability estimates of clinical deterioration, $\mathbf{\hat{y}}_\text{COVID-GMIC}^t, \mathbf{\hat{y}}_\text{COVID-GBM}^t \in [0,1]$.

In order to combine the predictions from COVID-GMIC and COVID-GBM, we employ the technique of model ensembling~\cite{dietterich2000ensemble}. Specifically, for each example, we compute a multi-modal prediction $\mathbf{\hat{y}}_\text{ENSEMBLE}$ as a linear combination of $\mathbf{\hat{y}}_{\text{COVID-GMIC}}$ and $\mathbf{\hat{y}}_{\text{COVID-GBM}}$: 
\begin{equation}
    \mathbf{\hat{y}}_{\text{ENSEMBLE}}=\lambda \mathbf{\hat{y}}_{\text{COVID-GMIC}}+(1-\lambda)\mathbf{\hat{y}}_{\text{COVID-GBM}},
\end{equation}
where $\lambda \in [0,1]$ is a hyperparameter. We selected the best $\lambda$ by optimizing the average of the AUC and PR AUC on the validation set. In Supplementary Figure~\ref{fig:top-features-xgboost-validation-ensemble}.b, we show the validation performance of $\mathbf{\hat{y}}_{\text{ENSEMBLE}}$ for varying $\lambda$.

\paragraph{Clinical variables model.}
The goal of the clinical variables model is to predict the risk of deterioration when the patient's vital signs are measured. Thus, each prediction was computed using a set of vital sign measurements, in addition to the patient's most recent laboratory test results, age, weight, and body mass index (BMI). The laboratory test results were represented as maximum and minimum statistics of all values collected within 12 hours prior to the time of the vital sign measurement. The feature sets of vital signs and laboratory tests were then processed using a gradient boosting model~\cite{ke2017lightgbm} which we refer to as COVID-GBM. For the final ensemble prediction, $\mathbf{\hat{y}}_{\text{ENSEMBLE}}$, we combined the COVID-GMIC prediction with the COVID-GBM prediction computed using the most recently collected clinical variables prior to the chest X-ray exam. In cases where there were no clinical variables collected prior to the chest X-ray, we performed a mean imputation of the predictions assigned to the validation set.

\paragraph{Chest X-ray image model.}
We process chest X-ray images using a deep convolutional neural network model, which we call COVID-GMIC, based on the GMIC model~\cite{shen2019globally,shen2020interpretable}. COVID-GMIC has two desirable properties. First, COVID-GMIC generates interpretable saliency maps that highlight regions in the X-ray images that correlate with clinical deterioration. Second, it possesses a local module that is able to utilize high-resolution information in a memory-efficient manner. This avoids aggressive downsampling of the input image, a technique that is commonly used on natural images~\cite{he2016deep,huang2017densely}, which may distort and blur informative visual patterns in chest X-ray images such as basilar opacities and pulmonary consolidation. In Supplementary Table~\ref{tab:reso}, we demonstrate that COVID-GMIC achieves comparable results to DenseNet-121, a neural network model that is not interpretable by design, but is commonly used for chest X-ray analysis~\cite{rajpurkar2017chexnet,allaouzi2019novel,liu2019sdfn,guan2020multi}.

The architecture of COVID-GMIC is schematically depicted in Figure~\ref{fig:system-overview}.b. COVID-GMIC processes an X-ray image $\mathbf{x} \in \mathbb{R}^{H,W}$ ($H$ and $W$ denote the height and width) in three steps. First, the global module helps COVID-GMIC learn an overall view of the X-ray image. Within this module, COVID-GMIC utilizes a global network $f_g$ to extract feature maps $\mathbf{h}_g \in \mathbb{R}^{h,w,n}$, where $h$, $w$, and $n$ denote the height, width, and number of channels of the feature maps. The resolution of the feature maps is chosen to be coarser than the resolution of the input image. For each time window $t \in \mathbb{T}_a$, we apply a $1{\times}1$ convolution layer with sigmoid activation to transform $\mathbf{h}_g$ into a saliency map $\mathbf{A}^t \in \mathbb{R}^{h,w}$ that highlights regions on the X-ray image which correlate with clinical deterioration.\footnote{For visualization purposes, we apply nearest neighbor interpolation to upsample the saliency maps to match the resolution of the original image.} Each element $\mathbf{A}^t_{i,j} \in [0,1]$ represents the contribution of the spatial location $(i,j)$ in predicting the onset of adverse events within time window $t$. In order to train $f_g$, we use an aggregation function $f_\text{agg}: \mathbb{R}^{h,w} \mapsto [0,1]$ to transform all saliency maps $\mathbf{A}^t$ for all time windows $t$ into classification predictions $\mathbf{\hat{y}}_\text{global}$:
\begin{equation}
    f_{\text{agg}}(\mathbf{A}^t) = \frac{1}{|H^+|}\sum_{(i,j) \in H^+} \mathbf{A}^t_{i,j},
\end{equation}
where $H^+$ denotes the set containing the locations of the $r\%$ largest values in $\mathbf{A}^t$, and $r$ is a hyperparameter.

The local module enables COVID-GMIC to selectively focus on a small set of informative regions. As shown in Figure~\ref{fig:system-overview}, COVID-GMIC utilizes the saliency maps, which contain the approximate locations of informative regions, to retrieve six image patches from the input X-ray image, which we call region-of-interest (ROI) patches. We refer the readers to supplementary note 6 for more details about the ROI retrieval algorithm. Figure~\ref{fig:gmic_vis} shows some examples of ROI patches. To utilize high-resolution information within each ROI patch $\mathbf{\tilde{x}} \in \mathbb{R}^{224, 224}$, COVID-GMIC applies a local network $f_l$, parameterized as a ResNet-18~\cite{he2016deep}, which produces a feature vector $\mathbf{\tilde{h}}_k \in \mathbb{R}^{512}$ from each ROI patch. The predictive value of each ROI patch might vary significantly. Therefore, we utilize the gated attention mechanism~\cite{ilse2018attention} to compute an attention score $\alpha_k \in [0,1]$ that indicates the relevance of each ROI patch $\mathbf{\tilde{x}}$ for the prediction task. To aggregate information from all ROI patches, we compute an attention-weighted representation:
\begin{equation}
    \mathbf{z} = \sum_{k=1}^6 \alpha_k \mathbf{\tilde{h}}_k.
\end{equation}
The representation $\mathbf{z}$ is then passed into a fully connected layer with sigmoid activation to generate a prediction $\mathbf{\hat{y}}_{\text{local}}$. We refer the readers to Shen et al.~\cite{shen2020interpretable} for further details.

The fusion module combines both global and local information to compute a final prediction. We apply global max pooling to $\mathbf{h}_g$, and concatenate it with $\mathbf{z}$ to combine information from both saliency maps and ROI patches. The concatenated representation is then fed into a fully connected layer with sigmoid activation to produce the final prediction $\mathbf{\hat{y}}_\text{fusion}$. 

In our experiments, we chose $H = W = 1024$. Supplementary Table~\ref{tab:reso} shows that COVID-GMIC achieves the best validation performance for this resolution.
We parameterize $f_g$ as a ResNet-18~\cite{he2016deep} which yields feature maps $\mathbf{h}^g$ with resolution $h=w=32$, and number of channels $n=512$. During training, we optimize the loss function:
\begin{equation}
    l(\mathbf{y}, \mathbf{\hat{y}}_\text{global}, \mathbf{\hat{y}}_\text{local}, \mathbf{\hat{y}}_\text{fusion}) = \frac{1}{|\mathbb{T}_a|} \sum_{t \in \mathbb{T}_a} \text{BCE}(\mathbf{y}^t,  \mathbf{\hat{y}}^t_\text{global}) + \text{BCE}(\mathbf{y}^t,  \mathbf{\hat{y}}^t_\text{local}) + \text{BCE}(\mathbf{y}^t,  \mathbf{\hat{y}}^t_\text{fusion}) + \beta |\mathbf{A}^t|,
\end{equation}
where BCE denotes binary cross-entropy and $\beta$ is a hyperparameter representing the relative weights on an $\ell_1$-norm regularization term that promotes sparsity of the saliency maps. During inference, we use $\mathbf{\hat{y}}_\text{fusion}$ as the final prediction generated by the model.

\paragraph{Estimation of deterioration risk curves.}
The deterioration risk curve (DRC) represents the evolution of the deterioration risk over time for each patient. Let $T$ denote the time of the first adverse event. The DRC is defined as a discretized curve that equals the probability $P(T \leq t_i)$ of the first adverse event occurring before time $t_i \in \{t_i |  1\leq i \leq 8 \}$, where $t_1=3$, $t_2=12$, $t_3= 24$, $t_4= 48$, $t_5= 72$, $t_6= 96$, $t_7= 144 $, $t_8= 192$ (all times are in hours). 

Following recent work on survival analysis via deep learning~\cite{gensheimer2018scalable}, we parameterize the DRC using a vector of conditional probabilities $\mathbf{\hat{p}} \in \mathbb{R}^{8}$. The $i^{th}$ entry of this vector, $\mathbf{\hat{p}}_i$, is equal to the conditional probability of the adverse event happening before time $t_i$ given that no adverse event occurred before time $t_{i-1}$, that is:\footnote{The parameters in our implementation are the complementary probabilities $\mathbf{\hat{q}}=1-\mathbf{\hat{p}}$, which is a mathematically equivalent parameterization. We also include an additional parameter to account for patients whose first adverse event occurs after 192 hours.} 
\begin{equation}
\mathbf{\hat{p}}_i =     \begin{cases} P(T \leq t_1) , \quad &  i = 1, \\
    P( T \leq t_i \, | \, T > t_{i-1}), & 2\leq i \leq 8.
    \end{cases}
\end{equation}
 
Given an estimate of $\mathbf{\hat{p}}$, the DRC can be computed applying the chain rule:

\begin{equation}
    \begin{split}
    \text{DRC}(t_i)& =P(T \leq t_i)\\
    & = 1-P(T > t_i)\\
    &= 1 - \prod_{j=1}^{i} P( T > t_j \, | \, T > t_{j-1})\\
    & = 1- \prod_{j=1}^{i} (1-\mathbf{\hat{p}}_j). 
    \end{split}
\end{equation}

We use the GMIC model to estimate the conditional probabilities $\mathbf{\hat{p}}$ from chest X-ray images. We refer to this model as COVID-GMIC-DRC. As explained in the previous section, the GMIC model has three different outputs corresponding to the global module, local module and fusion module. When estimating conditional probabilities for the eight time intervals, we denote these outputs by $\mathbf{\hat{p}}_\text{global}$, $\mathbf{\hat{p}}_\text{local}$, and $\mathbf{\hat{p}}_\text{fusion}$. During inference, we use the output of the fusion module, $\mathbf{\hat{p}}_\text{fusion}$, as the final prediction of the conditional-probability vector $\mathbf{\hat{p}}$. We use an input resolution of $H=W=512$ and parameterize $f_g$ as ResNet-34~\cite{he2016deep}. The resulting feature maps $\mathbf{h}_g$ have resolution $h=w=16$ and number of channels $n=512$. The results of an ablation study that evaluates the impact of input resolution and compares COVID-GMIC-DRC to a model based on the Densenet-121 architecture, are shown in the Supplementary Tables~\ref{tab:reso} and \ref{tab:survival_model_appendix}. 
During training, we minimize the following loss function defined on a single example:
\begin{equation}
    l(T, \mathbf{\hat{p}_\text{global},\hat{p}_\text{local},\hat{p}_\text{fusion}}) = l_s(T, \mathbf{\hat{p}}_\text{global}) +
    l_s(T,\mathbf{\hat{p}}_\text{local}) + l_s(T, \mathbf{\hat{p}}_\text{fusion}) + \sum_{m=0}^8 \beta|\mathbf{A}^m|,
\end{equation}
where $l_s$ is the negative log-likelihood of the conditional probabilities. For a patient who had an adverse event between $t_{i-1}$ and $t_i$ (where $t_0=0$), this negative log-likelihood is given by
\begin{equation}
    \begin{split}
        l_s(T,\mathbf{\hat{p}})& = -\ln   P(t_{i-1} \leq T \leq t_i)\\  & = -\ln  \prod_{j=1}^{i-1} P( T > t_j \, | \, T > t_{j-1})P( T \leq t_j \, | \, T > t_{i-1}) \\
      &= -\sum_{j=1}^{i-1}\ln(1-\mathbf{\hat{p}}_j) - \ln \mathbf{\hat{p}}_i \label{eqn:log-lik1}.
    \end{split}
\end{equation}

The framework can easily incorporate censored data corresponding to patients whose information is not available after a certain point. The negative log-likelihood corresponding to a patient, who has no information after $t_i$ and no adverse events before $t_i$, equals
\begin{equation}
    \begin{split}
        l_s(T,\mathbf{\hat{p}}) &= -\ln   P(T > t_i) \\  & = -\ln  \prod_{j=1}^{i} P( T > t_j \, | \, T > t_{j-1}) \\
      &= -\sum_{j=1}^{i}\ln(1-\mathbf{\hat{p}}_j) \label{eqn:log-lik2}.
    \end{split}
\end{equation}

Note that each $\mathbf{\hat{p}}_i$ is estimated only using patients that have data available up to $t_i$. The total negative log-likelihood of the training set is equal to the sum of the individual negative log-likelihoods corresponding to each patient, which makes it possible to perform minimization efficiently via stochastic gradient descent. In contrast, deep learning models for survival analysis based on Cox proportional hazards regression~\cite{cox1984analysis} require using the whole dataset to perform model updates~\cite{ching2018cox,katzman2018deepsurv,liang2020early}, which is computationally infeasible when processing large image datasets. 

\paragraph{Model training and selection.}

In this section, we discuss the experimental setup used for COVID-GMIC, COVID-GMIC-DRC, and COVID-GBM. The chest X-ray image models were implemented in PyTorch~\cite{NEURIPS2019_9015} and trained using NVIDIA Tesla V100 GPUs. The clinical variables models were implemented using the Python library LightGBM~\cite{ke2017lightgbm}.

We initialized the weights of COVID-GMIC and COVID-GMIC-DRC by pretraining them on the ChestX-ray14 dataset~\cite{wang2017chestx} (Supplementary Table~\ref{tab:transfer} compares the performance of different initialization strategies). We used Adam~\cite{kingma2014adam} with a minibatch size of eight to train the models on our data. During the training and test stages, we applied a set of data transformations to the inputs in order to make the model more robust to rotation and spatial translation. During the test stage, we applied ten different augmentations to each image and used the average of their predictions in order to further improve performance. We did not apply any data augmentation during the validation stage since it introduces randomness, which can be confounding when determining whether or not validation performance is improving.

We optimized the hyperparameters using random search~\cite{bergstra2012random}. For COVID-GMIC, we searched for the learning rate $\eta \in 10^{[-6,-4]}$ on a logarithmic scale, the regularization hyperparameter $\beta \in 4 \times 10^{[-6,-3]}$ on a logarithmic scale, and the pooling threshold $r \in [0.2,0.8]$ on a linear scale. For COVID-GMIC-DRC, based on the preliminary experiments, we fixed the learning rate to $1.25$ $\times$ $10^{-4}$. We searched for the regularization hyperparameter, $\beta \in 10^{[-6,-4]}$ on a logarithmic scale, and the pooling threshold $r \in \{0.2,0.5,0.8\}$. For COVID-GBM, we searched for the learning rate $\eta \in 10^{[-2,-1]}$ on a logarithmic scale, the number of estimators $e \in 10^{[2,3]}$ on a logarithmic scale, and the number of leaves $l \in [5,15]$ on a linear scale. For each hyperparameter configuration, we performed Monte Carlo cross-validation~\cite{xu2001monte} (we sampled $80\%$ of the data for training and $20\%$ of the data was used for validation). We performed cross-validation using three different random splits for each hyperparameter configuration. We then selected the top three hyperparameter configurations based on the average validation performance across the three splits. Finally, we combined the nine models from the top three hyperparameter configurations by averaging their predictions on the held-out test set to evaluate the performance. This procedure is formally described in Supplementary Algorithm~\ref{alg:model_selection}.

\paragraph{Design of the reader study}

The reader study consists of 200 frontal chest X-ray exams from the test set. We selected one exam per patient to increase the diversity of exams. We used stratified sampling to ensure that a sufficient number of exams in the study corresponded to the least common outcome (patients with adverse outcomes in the next 24 hours). In more detail, we oversampled exams of patients who developed an adverse event by sampling the first 100 exams only from patients from the test set that had an adverse outcome within the first 96 hours. The remaining 100 exams came from the remaining patients in the test set. The radiologists were asked to assign the overall probability of deterioration to each scan across all time windows of evaluation.

\section*{Acknowledgements}

The authors would like to thank Mario Videna, Abdul Khaja and Michael Constantino for supporting our computing environment, Philip P. Rodenbough (the NYUAD Writing Center) and Catriona C. Geras for revising the manuscript, and Boyang Yu, Jimin Tan, Kyunghyun Cho and Matthew Muckley for useful discussions. We also gratefully acknowledge the support of Nvidia Corporation with the donation of some of the GPUs used in this research. This work was supported in part by grants from the National Institutes of Health (P41EB017183, R01LM013316), the National Science Foundation (HDR-1922658, HDR-1940097), and NYU Abu Dhabi.

\section*{Author Contributions}

FES, YS, NW, AK, JP and TM designed and conducted the experiments with neural networks. FES, NW, JP, SJ, TM and JW built the data preprocessing pipeline. FES, NR and BZ designed the clinical variables model. SJ conducted the reader study and analyzed the data. SD and MC conducted literature search. YL, DW, BZ and YA collected the data. DK, LA and WM analyzed the results from a clinical perspective. YA, CFG and KJG supervised the execution of all elements of the project. All authors provided critical feedback and helped shape the manuscript.

\section*{Competing Interests}

The authors declare no competing interests.

\section*{Data Availability}
The ImageNet dataset is available at \href{http://www.image-net.org/}{\url{http://www.image-net.org/}}. The ChestX-ray8 dataset is available at \href{https://nihcc.app.box.com/v/ChestXray-NIHCC}{\url{https://nihcc.app.box.com/v/ChestXray-NIHCC}}. The COVID-19 X-ray images and associated clinical variables from NYU Langone Health are not publicly available, but we provide sample patients in our source code repository. 

\section*{Code Availability}
The code of the models in this study, along with their trained weights, are available at \href{https://github.com/nyukat/COVID-19_prognosis}{\url{https://github.com/nyukat/COVID-19_prognosis}}.

\bibliographystyle{customize_naturemag}

\newpage

\renewcommand\thefigure{\arabic{figure}}  
\renewcommand\thetable{\arabic{table}}

\section*{Supplementary Information} \label{sec:supplementary_information}
\setcounter{figure}{0}   
\setcounter{table}{0}    

\section*{Supplementary Note 1: Image preprocessing}
In Supplementary Figure~\ref{fig:preprocessing}.a, we show the heights and widths of the images prior to data augmentation. In Supplementary Figure~\ref{fig:preprocessing}.b, we show an example of a raw image and the final image after applying the preprocessing steps in Figure~\ref{fig:preprocessing}.c.

\begin{figure}[ht]
\renewcommand\figurename{Supplementary Figure}
     \centering
     \includegraphics[width=\textwidth]{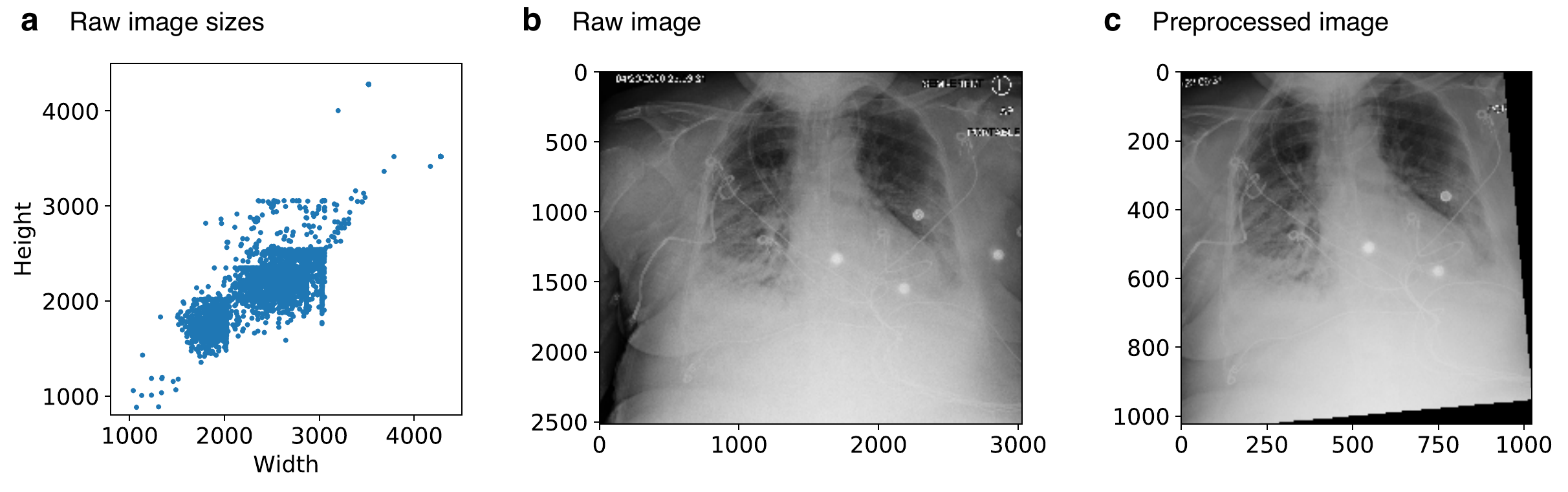}
     \caption{\small (a) Heights and widths (in pixels) of images prior to data augmentation. (b) An example raw image. (c) To ensure that the inputs to the model have a consistent size, we perform center cropping and rescaling. In addition, we apply random horizontal flipping, rotation, and translation to augment the training dataset.}
     \label{fig:preprocessing}
\end{figure}

\newpage

\section*{Supplementary Note 2: Clinical variables modeling}

The average importance of the top ten features computed by the COVID-GBM models are shown in Supplementary Figure~\ref{fig:top-features-xgboost-validation-ensemble}.a. The importance of a feature is measured by the numbers of times the feature is used to split the data across all trees in a single COVID-GBM model. Age is amongst the top ten features across all time windows, which is consistent with existing findings that mortality is more common amongst elderly COVID-19 patients than younger patients~\cite{liu2020clinical}. The inclusion of the vital sign variables, amongst the top ten features across all models, is also aligned with existing research suggesting that they are strong indicators of deterioration~\cite{news}.

\begin{figure}[H]
\renewcommand\figurename{Supplementary Figure}
    \centering
    \includegraphics[width=1\textwidth]{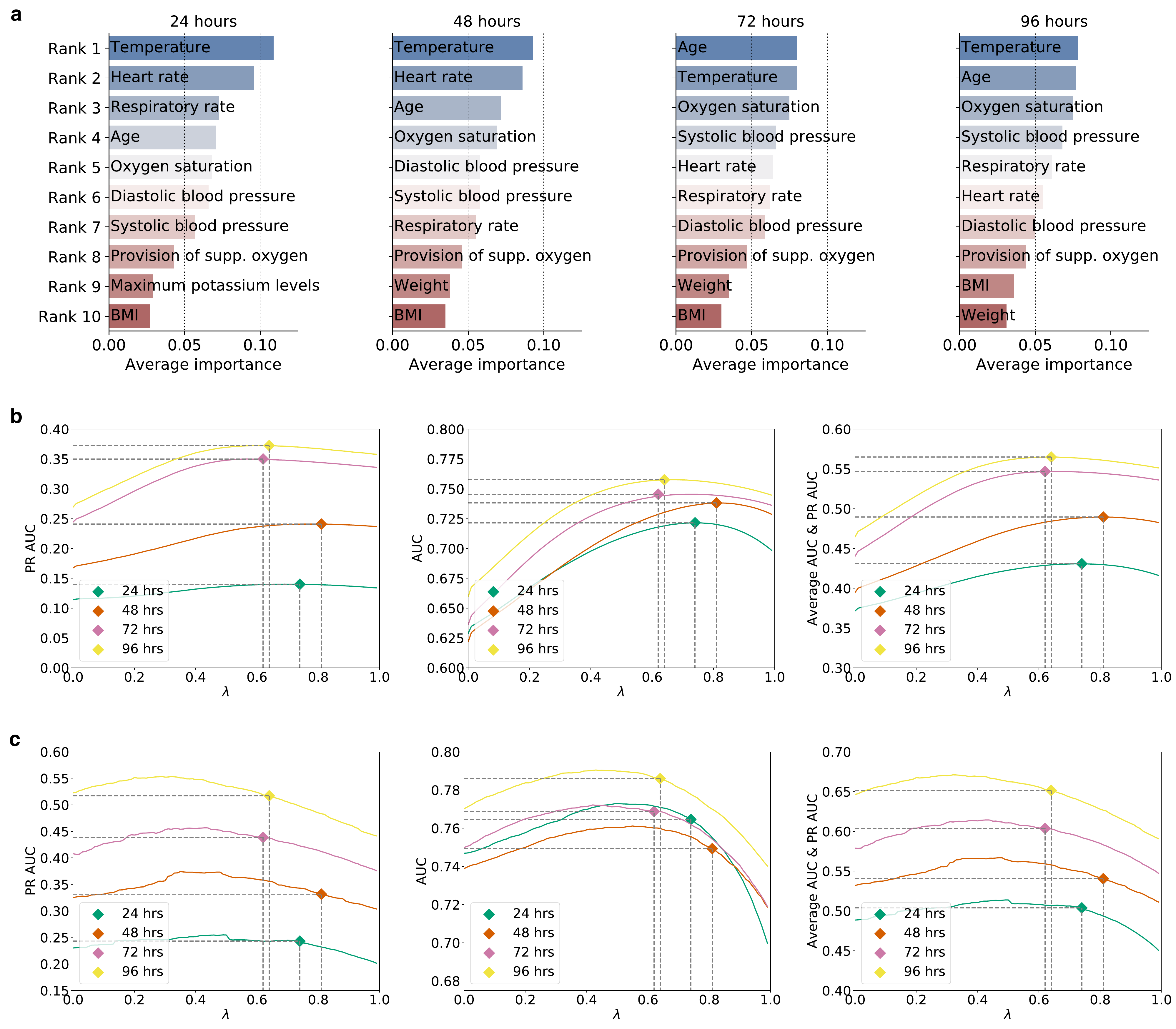}
    \caption{\small \textbf{Additional results for COVID-GBM and the ensemble of COVID-GBM and COVID-GMIC.} \textbf{a}, The average importance of the top ten features computed by the nine COVID-GBM ensemble models for 24, 48, 72, and 96 hours. The importance of a feature is measured by the numbers of times the feature is used to split the data across all trees in a model. \textbf{b}, The effect of varying $\lambda$, the weight on the COVID-GMIC prediction, in combining the predictions of COVID-GMIC and COVID-GBM when using AUC, PR AUC and the average AUC and PR AUC on the validation set. For the average AUC and PR AUC, the optimal $\lambda$ was 0.74 for 24 hours, 0.81 for 48 hours, 0.62 for 72 hours, and 0.64 for 96 hours. \textbf{c}, the optimal values of $\lambda$ selected through the validation set in \textbf{b} are shown for the test set.}
    \label{fig:top-features-xgboost-validation-ensemble}
\end{figure}

\newpage

\section*{Supplementary Note 3: Ablation studies}

\paragraph{DenseNet-121-based models.}
DenseNet~\cite{huang2017densely} is a deep neural network architecture which consists of dense blocks in which layers are directly connected to every other layer in a feed-forward fashion. It achieves strong performance on benchmark natural images dataset, such as CIFAR10/100~\cite{krizhevsky2009learning} and ILSVRC 2012 (ImageNet) dataset~\cite{imagenet_cvpr09} while being computationally efficient. Here we compare COVID-GMIC to a specific variant of DenseNet, DenseNet-121, which has been applied to process chest X-ray images in the literature~\cite{rajpurkar2017chexnet,allaouzi2019novel,liu2019sdfn,guan2020multi}. 

The model assumes an input size of $224{\times}224$. We applied DenseNet-121-based models to predict deterioration and also to compute deterioration risk curves. We initialized the models with weights pretrained on the ChestX-ray14 dataset~\cite{wang2017chestx}, provided at \hyperlink{"https://github.com/arnoweng/CheXNet"}{https://github.com/arnoweng/CheXNet}. We used weight decay in the optimizer. To perform hyperparameter search, we sampled the learning rate and the rate of weight decay per step uniformly on a logarithmic scale between $10^{[-6, -1]}$ and $10^{[-6, -3]}$.

\begin{figure}[H]
\renewcommand\figurename{Supplementary Figure}
    \includegraphics[width=\textwidth]{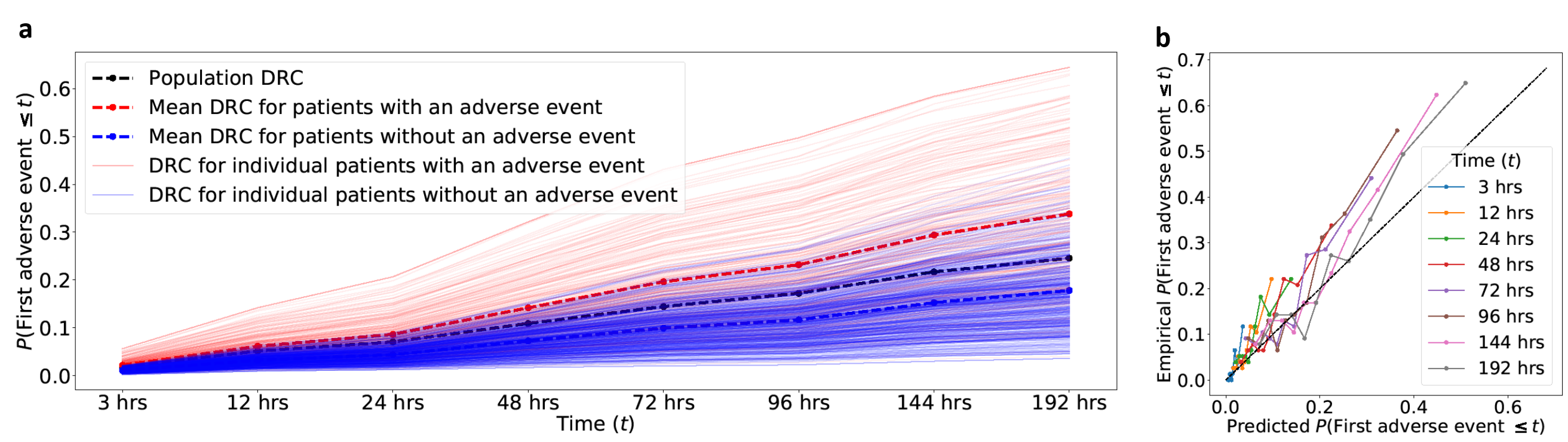}
    \caption{\small \textbf{Deterioration risk curves (DRCs) and reliability plot for DenseNet-121.} Compare to Figure~\ref{fig:gmic_survival_model}, which shows analogous graphs for COVID-GMIC-DRC. \textbf{a}, DRCs generated by DenseNet-121 model for patients in the test set with (faded red lines) and without adverse events (faded blue lines). The mean DRC for patients with adverse events (red dashed line) is higher than the DRC for patients without adverse events (blue dashed line) at all times. The graph also includes the ground-truth population DRC (black dashed line) computed from the test data. \textbf{b}, Reliability plot of the DRCs generated by DenseNet-121 model for patients in the test set. The empirical probabilities are computed by dividing the patients into deciles according to the value of the DRC at each time $t$. The empirical probability equals the fraction of patients in each decile that suffered adverse events up to $t$. This is plotted against the predicted probability, which equals the mean DRC of the patients in the decile. The diagram shows that these values are similar across the different values of $t$, and hence the model is well-calibrated (for comparison, perfect calibration would correspond to the diagonal black dashed line).}
    \label{fig:dn_survival_model}
\end{figure}

For adverse event prediction, the DenseNet-121 based model yielded test AUCs of 0.687 (95\% CI: 0.621 - 0.749), 0.709 (95\% CI: 0.653 - 0.757), 0.710 (95\% CI: 0.660 - 0.763), and 0.736 (95\% CI: 0.691 - 0.782), and PRAUCs of 0.216 (95\% CI: 0.155 - 0.317), 0.315 (95\% CI: 0.239 - 0.419), 0.373 (95\% CI: 0.300 - 0464), and 0.454 (95\% CI: 0.384 - 0.542) for 24, 48, 72, and 96 hours. The deterioration risk curves produced by the DenseNet-121 based models and the corresponding reliability plot are presented in Figure~\ref{fig:dn_survival_model}.

\paragraph{Impact of input image resolution.}

Prior work on deep learning for medical images~\cite{geras2017high} report that using high resolution input images can improve  performance. In this section, we analyze the impact of image resolution on our tasks of interest. We consider the following image sizes: $128{\times}128$, $256{\times}256$, $512{\times}512$, and $1024{\times}1024$. We pretrain all models on the ChestX-ray14 dataset~\citep{wang2017chestx} and then fine-tune them on our dataset. Results on the test set are reported in Supplementary Table~\ref{tab:reso}. 

The DenseNet-121 based model achieves the best AUCs when using an image size of $256\times256$, and the best concordance index for $512{\times}512$. Further increasing the resolution does not improve performance. COVID-GMIC achieves the best performance for the highest input image resolution of $1024{\times}1024$, while achieving the best concordance index for $512{\times}512$. While a further increase in performance may be possible, we did not consider any larger image sizes resolutions because the computational cost would become prohibitively high.

\begin{table}[H]
\renewcommand\tablename{Supplementary Table}
\centering
\caption{\small Model performance with 95\% confidence intervals when using input images of sizes of $128{\times}128$, $256{\times}256$, $512{\times}512$, and $1024{\times}1024$. For COVID-GMIC, we started with a size of $256{\times}256$ since an image with resolution of $128{\times}128$ pixels results in saliency maps that are too small to generate meaningful ROI patches. We report AUCs for predicting the risk of deterioration within 24, 48, 72, and 96 hours. When evaluating the deterioration risk curves, we report the concordance index with a reference time of 96 hours, as well as the average of the index over all possible reference times (3, 12, 24, 48, 72, 96, 144, and 192 hours).}
\resizebox{.995\textwidth}{!}{\begin{tabular}{@{}lllllllll@{}}  
    \toprule 
     && \multicolumn{4}{c}{AUC / PR AUC} &\phantom{a}& \multicolumn{2}{c}{Concordance index}\\
    \cmidrule{3-6} \cmidrule{8-9}
     &&\multicolumn{1}{c}{24 hours} &\multicolumn{1}{c}{48 hours} & \multicolumn{1}{c}{72 hours} & \multicolumn{1}{c}{96 hours} && \multicolumn{1}{c}{96 hours} & \multicolumn{1}{c}{Average}\\
   \midrule 
     DenseNet-121 & $128{\times}128$ & 0.663 (0.602, 0.733) / & 0.688 (0.633, 0.749) / & 0.700 (0.649, 0.753) / & 0.728 (0.685, 0.781) / && 0.700 (0.667, 0.734)  & 0.700 (0.672, 0.736) \\
      &  & 0.214 (0.119, 0.284)  & 0.300 (0.198, 0.376)  & 0.370 (0.279, 0.448)  & 0.453 (0.364, 0.533)  &&& \\
     
      & $256{\times}256$ & \textbf{0.698} (0.632, 0.763) / & \textbf{0.721} (0.668, 0.778) / & \textbf{0.719} (0.670, 0.773) / & \textbf{0.748} (0.701, 0.795) / && 0.701 (0.666, 0.738)  & 0.698 (0.663, 0.734) \\
      & & \textbf{0.218} (0.153, 0.310) & 0.310 (0.207, 0.382)  & \textbf{0.390} (0.318, 0.486) & \textbf{0.469} (0.392, 0.562)&&&\\
      
      &$512{\times}512$ & 0.682 (0.617, 0.749) / & 0.710 (0.658, 0.764) / & 0.709 (0.656, 0.764) / & 0.732 (0.686, 0.780) / && \textbf{0.705} (0.673,0.739) & \textbf{0.701} (0.669,0.735)\\
      &&0.208 (0.111, 0.267)  & \textbf{0.318} (0.238, 0.422) & 0.383 (0.286, 0.459)  & 0.441 (0.353, 0.516) &&& \\

      &$1024{\times}1024$ & 0.680 (0.619, 0.742) / & 0.709 (0.657, 0.763) / & 0.716 (0.666, 0.766) / & 0.739 (0.694, 0.787) / &  & 0.701 (0.668, 0.734)  & 0.696 (0.664, 0.729) \\
      & & 0.180 (0.101, 0.230)  & 0.278 (0.185, 0.344)  & 0.369 (0.269, 0.442)  & 0.441 (0.353, 0.516) &&&\\

      \midrule 
     COVID-GMIC & $256{\times}256$ & 0.664 (0.593, 0.734) / & 0.688 (0.630, 0.747) / & 0.699 (0.651, 0.750) / & 0.728 (0.684, 0.774) / &  & 0.712 (0.679, 0.744)  & 0.707 (0.675, 0.741) \\
     &  & 0.202 (0.101, 0.260)  & 0.263 (0.172, 0.326)  & 0.342 (0.253, 0.414)  & 0.424 (0.343, 0.492) &&&\\

     & $512{\times}512$ & \textbf{0.700} (0.635, 0.765) / & 0.714 (0.661, 0.769) / & 0.714 (0.671, 0.766) / & 0.733 (0.690, 0.780) / &  & \textbf{0.713} (0.679,0.748) & \textbf{0.708} (0.675, 0.742)\\
     && \textbf{0.210} (0.154, 0.298) & 0.300 (0.205, 0.370)  & \textbf{0.389} (0.314, 0.481) & 0.443 (0.354, 0.515) &&&\\
     
    & $1024{\times}1024$ & 0.695 (0.630, 0.763) / & \textbf{0.716} (0.661, 0.767) / & \textbf{0.717} (0.663, 0.764) / & \textbf{0.738} (0.692, 0.780) / &  & 0.686 (0.650, 0.720)  & 0.685 (0.648, 0.717)\\
    && 0.200 (0.121, 0.258)  & \textbf{0.302} (0.230, 0.394) & 0.374 (0.289, 0.447)  & \textbf{0.439} (0.368, 0.522) &&&\\

    \bottomrule \\
    \end{tabular}}

 \label{tab:reso}
\end{table}

\paragraph{Impact of different transfer learning strategies.}
In data-scarce applications, it is crucial to pretrain deep neural networks on a related task for which a large dataset is available, prior to fine-tuning on the task of interest~\cite{pan2009survey, yosinski2014transferable}. Given the relatively small number of COVID-19 positive cases in our dataset, we investigate the impact of different weight initialization strategies on our results. Specifically, we compare three strategies: 1) initialization by He et al.~\cite{he2015delving}, 2) initialization with weights from models trained on natural images (ImageNet~\cite{imagenet_cvpr09}), and 3) initialization with weights from models trained on chest X-ray images (ChestX-ray14 dataset~\citep{wang2017chestx}). We apply the initialization procedure to all layers except the last fully connected layer, which is always initialized randomly. We then fine-tune the entire network on our COVID-19 task.

Based on results shown in Supplementary Table~\ref{tab:transfer}, fine-tuning the network from weights pretrained on the ChestX-ray14 dataset is the most effective strategy for COVID-GMIC. This dataset contains over 100,000 chest X-ray images from more than 30,000 patients, including many with advanced lung disease. The images are paired with labels representing fourteen common thoracic observations: atelectasis, cardiomegaly, effusion, infiltration, mass, nodule, pneumonia, pneumothorax, consolidation, edema, emphysema, fibrosis, pleural thickening, and hernia. By pretraining a model to detect these conditions, we hypothesize that the model learns a representation that is useful for our downstream task of COVID-19 prognosis.

\begin{table}[H]
\renewcommand\tablename{Supplementary Table}
\centering
\caption{\small Model performance with 95\% confidence intervals across three different initialization strategies: random initialization, initialization with the weights of the model pretrained on ImageNet~\cite{imagenet_cvpr09} and initialization with the weights of the model pretrained model on the ChestX-ray14 dataset~\cite{wang2017chestx}. We report AUCs for each time window in the outcome classification task. When evaluating the deterioration risk curves, we report the concordance index with a reference time of 96 hours, as well as the average of the index over all discretized times (3, 12, 24, 48, 72, 96, 144, and 192 hours).}
\resizebox{.995\textwidth}{!}{\begin{tabular}{@{}lllllllll@{}}  
    \toprule 
     & & \multicolumn{4}{c}{AUC / PR AUC}& \phantom{a}& \multicolumn{2}{c}{Concordance index}\\
    \cmidrule{3-6} \cmidrule{8-9}
      &&\multicolumn{1}{c}{24 hours} & \multicolumn{1}{c}{48 hours} & \multicolumn{1}{c}{72 hours} & \multicolumn{1}{c}{96 hours} && \multicolumn{1}{c}{96 hours} & \multicolumn{1}{c}{Average}\\
   \midrule 
     DenseNet-121 & Random & 0.687 (0.625, 0.753) / & 0.699 (0.648, 0.754) / & 0.693 (0.642, 0.747) / & 0.705 (0.660, 0.752) / && 0.649 (0.614, 0.686)  & 0.648 (0.613, 0.685)\\
     & & 0.178 (0.105, 0.222)  & 0.258 (0.177, 0.315)  & 0.326 (0.236, 0.388)  & 0.386 (0.298, 0.449) &&  & \\
      & ImageNet & \textbf{0.701} (0.639, 0.761) / & \textbf{0.722} (0.668, 0.776) / & \textbf{0.719} (0.670, 0.772) / & \textbf{0.745} (0.701, 0.789) / &  & 0.686 (0.652, 0.720)  & 0.683 (0.651, 0.715)
      \\&& 0.206 (0.117, 0.260)  & 0.299 (0.197, 0.366)  & 0.365 (0.264, 0.436)  & 0.444 (0.349, 0.513) &&  & \\
      & ChestX-ray14 & 0.687 (0.616, 0.755) / & 0.709 (0.651, 0.765) / & 0.710 (0.657, 0.760) / & 0.736 (0.690, 0.781) / &  & \textbf{0.705} (0.673,0.739) & \textbf{0.701} (0.669,0.735)\\
      && \textbf{0.216} (0.155, 0.317) & \textbf{0.315} (0.239, 0.419) & \textbf{0.373} (0.300, 0.464) & \textbf{0.454} (0.384, 0.542) && &
      \\
     \midrule 
    COVID-GMIC & Random & 0.675 (0.609, 0.743) / & 0.671 (0.614, 0.725) / & 0.686 (0.640, 0.732) / & 0.708 (0.668, 0.752) / &  & 0.643 (0.606, 0.678)  & 0.640 (0.604, 0.673)\\
     &&  0.174 (0.101, 0.223)  & 0.227 (0.146, 0.277)  & 0.290 (0.214, 0.345)  & 0.352 (0.276, 0.410) &&  & 
     \\
     & ImageNet & 0.694 (0.635, 0.757) / & 0.709 (0.657, 0.761) / & extbf{0.724} (0.673, 0.769) / & 0.737 (0.696, 0.782) / && 0.684 (0.652, 0.717)  & 0.680 (0.649, 0.711)\\
     &&  0.195 (0.110, 0.252)  & 0.258 (0.165, 0.319)  & 0.347 (0.263, 0.416)  & 0.433 (0.354, 0.506)&& &
     \\
     & ChestX-ray14 & \textbf{0.695} (0.626, 0.757) / & \textbf{0.716} (0.659, 0.768) / & 0.717 (0.672, 0.769) / & \textbf{0.738} (0.690, 0.783) / &  & \textbf{0.713} (0.679,0.748) & \textbf{0.708} (0.675,0.742)\\
     && \textbf{0.200} (0.142, 0.283) & \textbf{0.302} (0.228, 0.400) & \textbf{0.374} (0.302, 0.463) & \textbf{0.439} (0.368, 0.532) &&&\\
    \bottomrule \\
    \end{tabular}}
 \label{tab:transfer}
\end{table}

\paragraph{Impact of training set size.}
We evaluated the impact of the sample size used for training our machine learning models. Specifically, we evaluated our models on a subset of the training data, obtained by randomly sampling 12.5\%, 25\%, and 50\% of the exams. Table~\ref{tab:trainingsize} presents the AUCs and PR AUCs and the concordance indices achieved on the test set. It is evident that the performance of COVID-GMIC and COVID-GBM improve when increasing the number of images and clinical variables used for training, which highlights the importance of using a large dataset.

\begin{table}[H]
\renewcommand\tablename{Supplementary Table}
\centering
\caption{\small Model performance with 95\% confidence intervals when using 12.5\%, 25\%, 50\%, and 100\% of the training data. We report AUCs for each time window in the adverse event prediction task. When evaluating the deterioration risk curves, we report the concordance index with a reference time of 96 hours, as well as the average of the index over all discretized times (3, 12, 24, 48, 72, 96, 144, and 192 hours).}
\resizebox{.995\textwidth}{!}{\begin{tabular}{@{}lllllllll@{}}  
    \toprule 
    & & \multicolumn{4}{c}{AUC / PR AUC} & \phantom{a}& \multicolumn{2}{c}{Concordance index}\\
    \cmidrule{3-6} \cmidrule{8-9}
    & & \multicolumn{1}{c}{24 hours} & \multicolumn{1}{c}{48 hours} & \multicolumn{1}{c}{72 hours} & \multicolumn{1}{c}{96 hours} & & \multicolumn{1}{c}{96 hours} & \multicolumn{1}{c}{Average} \\
   \midrule 
     DenseNet-121 & 12.5\% & 0.608 (0.530, 0.678) / & 0.653 (0.594, 0.711) / & 0.672 (0.617, 0.722) / & 0.703 (0.654, 0.749) / && 0.675 (0.640, 0.708)  & 0.670 (0.636, 0.703)\\
      &  &  0.182 (0.094, 0.241)  & 0.265 (0.177, 0.332)  & 0.336 (0.248, 0.401)  & 0.415 (0.330, 0.486)& & &\\
     
     & 25\% & 0.638 (0.570, 0.708) / & 0.678 (0.621, 0.737) / & 0.682 (0.628, 0.734) / & 0.711 (0.662, 0.758) / && 0.676 (0.641, 0.709) & 0.671 (0.637, 0.704)\\
     &  & 0.174 (0.090, 0.227)  & 0.266 (0.170, 0.327)  & 0.327 (0.239, 0.393)  & 0.408 (0.321, 0.475) && & \\

     & 50\% & 0.672 (0.605, 0.737) / & 0.699 (0.644, 0.752) / & 0.698 (0.646, 0.747) / & 0.725 (0.679, 0.769) / &  & 0.694 (0.660, 0.728)  & 0.691 (0.657, 0.725)\\
     & &0.214 (0.109, 0.278)  & 0.303 (0.209, 0.373)  & 0.351 (0.265, 0.417)  & 0.433 (0.349, 0.501) & & & \\

     & 100\% & \textbf{0.687} (0.621, 0.753) / & \textbf{0.709} (0.654, 0.763) / & \textbf{0.710} (0.658, 0.761) / & \textbf{0.736} (0.689, 0.781) / &  & \textbf{0.705} (0.673,0.739) & \textbf{0.701} (0.669,0.735)\\
     &  & \textbf{0.216} (0.154, 0.317) & \textbf{0.315} (0.239, 0.417) & \textbf{0.373} (0.298, 0.475) & \textbf{0.454} (0.377, 0.552) &&  & \\

     \midrule
     COVID-GMIC & 12.5\% & 0.640 (0.577, 0.703) / & 0.672 (0.621, 0.726) / & 0.677 (0.631, 0.728)  & 0.695 (0.652, 0.738) / &  & 0.673 (0.640, 0.706)  & 0.668 (0.635, 0.701)\\
     &  &  0.145 (0.084, 0.180)  & 0.231 (0.146, 0.283)  & 0.318 (0.230, 0.387)  & 0.384 (0.294, 0.449) &&& \\
     
     & 25\% & 0.661 (0.598, 0.724) / & 0.672 (0.616, 0.726) / & 0.677 (0.627, 0.723) / & 0.693 (0.649, 0.738) / &  & 0.689 (0.655, 0.723)  & 0.680 (0.646, 0.714)\\
     &  &  0.177 (0.091, 0.229)  & 0.254 (0.162, 0.312)  & 0.327 (0.238, 0.388)  & 0.395 (0.313, 0.461) &&& \\

     & 50\% &0.646 (0.576, 0.715) / & 0.681 (0.624, 0.740) / & 0.687 (0.635, 0.742) / & 0.716 (0.669, 0.764) / && 0.699 (0.664, 0.733)  & 0.690 (0.657, 0.722)\\
     &  &0.164 (0.090, 0.212)  & 0.266 (0.172, 0.333)  & 0.351 (0.257, 0.428)  & 0.424 (0.332, 0.502) &&&\\

     & 100\% & \textbf{0.695} (0.626, 0.753) / & \textbf{0.716} (0.663, 0.769) / & \textbf{0.717} (0.667, 0.767) / & \textbf{0.738} (0.693, 0.782) / && \textbf{0.713} (0.679,0.748) & \textbf{0.708} (0.675,0.742)\\
     & &  \textbf{0.200} (0.142, 0.276) & \textbf{0.302} (0.230, 0.395) & \textbf{0.374} (0.297, 0.461) & \textbf{0.439} (0.363, 0.521) &&&\\

     \midrule
     COVID-GBM & 12.5\% & 0.674 (0.609, 0.736) / & 0.699 (0.647, 0.753) / & 0.710 (0.666, 0.761) / & 0.708 (0.663, 0.755) /  &  &  &  \\
     &  &  0.262 (0.153, 0.344)  & 0.297 (0.199, 0.366)  & 0.395 (0.310, 0.472)  & 0.439 (0.361, 0.516) &  &  &  \\
     
     & 25\% & 0.688 (0.628, 0.740) / & 0.716 (0.666, 0.765) / & 0.733 (0.689, 0.778) / & 0.739 (0.695, 0.784) /  &  &  &  \\
     &  & 0.175 (0.102, 0.220)  & 0.319 (0.227, 0.401)  & 0.385 (0.304, 0.461)  & 0.476 (0.402, 0.545) &  &  &  \\

     & 50\% & 0.743 (0.699, 0.796) / & \textbf{0.752} (0.702, 0.797) / & 0.749 (0.706, 0.795) / & 0.751 (0.711, 0.796) /  & &  & \\
     && 0.210 (0.119, 0.263)  & \textbf{0.325} (0.252, 0.425) & 0.418 (0.326, 0.495)  & 0.482 (0.396, 0.557) & &  & \\
     
     & 100\% & \textbf{0.747} (0.692, 0.798) / & 0.739 (0.687, 0.793) / & \textbf{0.750} (0.704, 0.794) / & \textbf{0.770} (0.728, 0.811) /  &&  & \\
     &  & \textbf{0.230} (0.167, 0.322) & \textbf{0.325} (0.253, 0.425) & \textbf{0.408} (0.334, 0.502) & \textbf{0.523} (0.439, 0.611) &&  & \\

    \bottomrule \\
    \end{tabular}}
 \label{tab:trainingsize}
\end{table}

\newpage

\section*{Supplementary Note 4: Additional results on the test sets}
We visualize the receiver operating characteristic (ROC) and precision-recall (PR) curves on the test set in Supplementary Figure~\ref{fig:all_peformance_task_1}. In \textbf{a}, we group the results based on the predictive models (COVID-GMIC, COVID-GBM, and the ensemble of both), while in \textbf{b}, we group the performances based on the time window of the task (i.e., 24, 48, 72, and 96 hours). In Supplementary Figure~\ref{fig:reader_study_roc_pr}, we visualize the ROC and PR curves on the test set considered in the reader study. 

\begin{figure}[H]
\renewcommand\figurename{Supplementary Figure}
    \centering
    \includegraphics[width=1\textwidth]{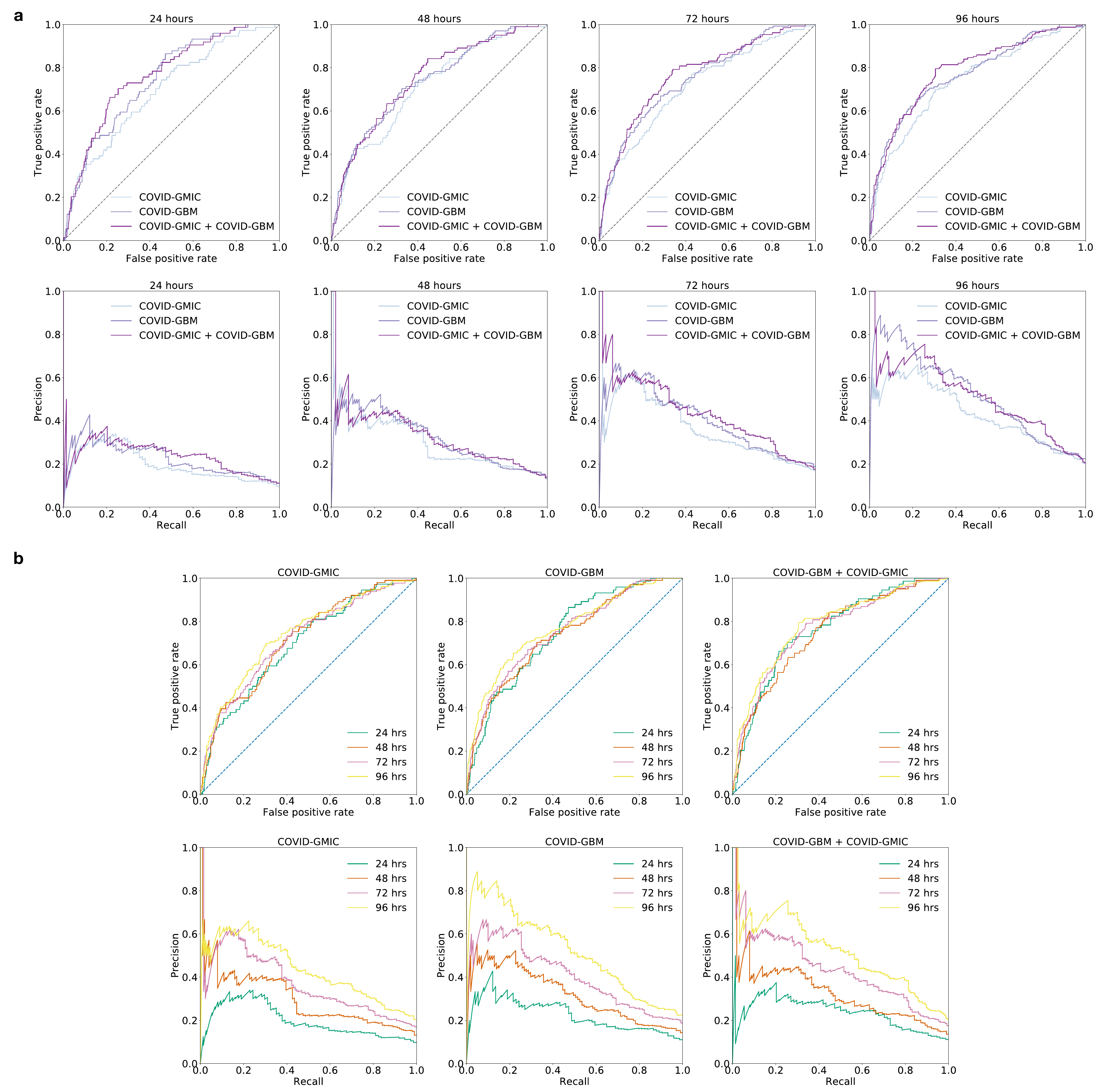}
    \caption{\small \textbf{Receiver operating characteristic (ROC) and Precision-Recall (PR) curves for predicting the onset of adverse events within 24, 48, 72, and 96 hours evaluated on the test set.} \textbf{a}, ROC and PR curves are grouped by predictive models. Ensembling COVID-GMIC and COVID-GBM improves performance in almost all cases. \textbf{b}, ROC and PR curves are grouped by time window of the task. The AUC and PR AUC improve as the length of the time window increases, which is consistence across models. Numerical values of AUCs and PR AUCs can be found in Table~\ref{tab:overall-performance}.}
    \label{fig:all_peformance_task_1}
\end{figure}

\begin{figure}[H]
\renewcommand\figurename{Supplementary Figure}
\centering
     \includegraphics[width=1\textwidth]{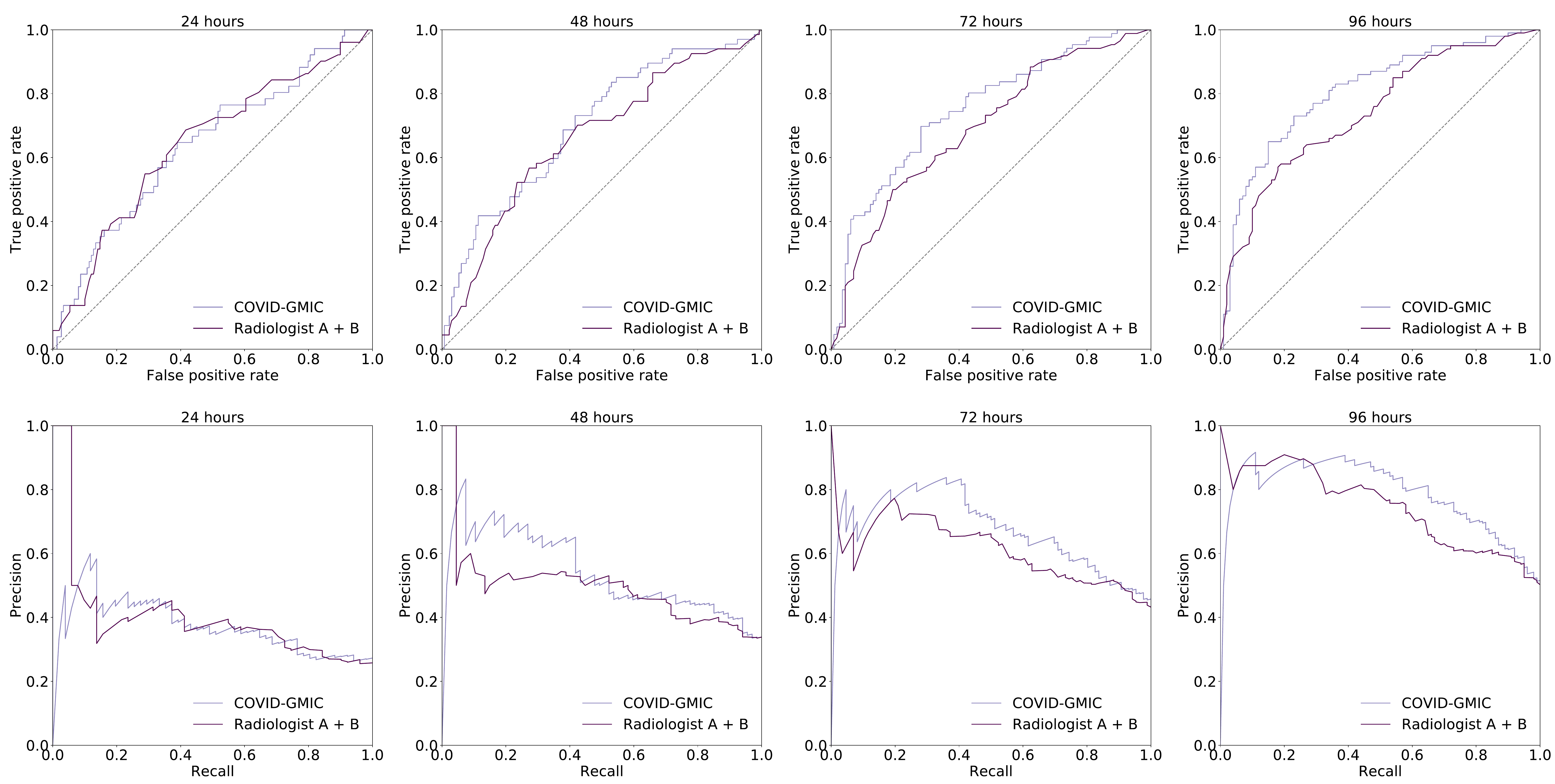}
    \caption{\small Test set ROC (top) and PR (bottom) curves of COVID-GMIC and the radiologists for predicting the risk of deterioration over 24, 48, 72, and 96 hours. These results suggest that COVID-GMIC performs comparably to the radiologists. Numerical values of AUCs and PR AUCs can be found in Table~\ref{tab:overall-performance}.}
    \label{fig:reader_study_roc_pr}
\end{figure}

In Supplementary Table~\ref{tab:survival_model_appendix}, we show the concordance index results across all time intervals for the best DenseNet-121 and COVID-GMIC-DRC models.

\begin{table}[H]
\renewcommand\tablename{Supplementary Table}
\centering
\caption{\small Concordance index (with 95\% confidence intervals) of the DRC curves generated by the best DenseNet-121 and COVID-GMIC-DRC models. Both models use input images of size $512{\times}512$ and are pretrained on the ChestX-ray14 dataset~\cite{wang2017chestx}. The results shows that the concordance index does not change much with the choice of time reference.
}
\resizebox{.995\textwidth}{!}{\begin{tabular}{@{}cccccccccc@{}}  
    \toprule 
     &\multicolumn{9}{c}{Concordance index}\\
    \cmidrule{2-10}
     Time (in hours) & 3 & 12 & 24 & 48 & 72 & 96 & 144 & 192 & Ave.\\
   \midrule 
     DenseNet-121 & 0.681 & 0.694 & 0.701  & 0.702  &0.703 & 0.705 &0.706 &0.705 &0.701 \\
   & (0.647, 0.714) & (0.658, 0.727) & (0.666, 0.735) & (0.667, 0.735) & (0.668, 0.734) & (0.671, 0.737) & (0.672, 0.739) & (0.67, 0.737) & (0.667, 0.733)\\
      \midrule 
     COVID-GMIC-DRC & 0.692 & 0.698 & 0.706 & 0.710 & 0.712 & 0.713 & 0.716 & 0.715 & \textbf{0.708} \\
   & (0.661, 0.734) & (0.664, 0.736) & (0.672, 0.74) & (0.677, 0.746) & (0.676, 0.745) & (0.678, 0.747) & (0.681, 0.748) & (0.68, 0.748) & (0.674, 0.741)\\
    \bottomrule \\
    \end{tabular}}
 \label{tab:survival_model_appendix}
\end{table}

\newpage
\section*{Supplementary Note 5: Model selection}
We describe our model selection procedure used throughout the paper in Algorithm \ref{alg:model_selection}. For the ablation study in Supplementary Table \ref{tab:trainingsize}, we control the size of the dataset by setting the parameter $u$ to $12.5$, $25$ and $50$. Specifically, in that experiment, we randomly sampled $u\%$ of the training set $\mathcal{D}_t$ as the ``universe'' $\mathcal{U}$ that our model used for training and validation.

\renewcommand{\algorithmicrequire}{\textbf{Input:}}
\renewcommand{\algorithmicensure}{\textbf{Output:}}
\newcommand{\RETURN}{\textbf{return}}
\begin{algorithm}[htb!]
    \caption{Model selection}
    \label{alg:model_selection}
    \begin{algorithmic}[1]
        \REQUIRE training set $\mathcal{D}_t$, test set $\mathcal{D}_s$, universe fraction $u \in [0, 100]$, and a predictive model $\mathcal{M}$
        \ENSURE $a^*$ performance of $\mathcal{M}$ evaluated on $\mathcal{D}_s$
        \STATE{$\mathcal{U} = $ randomly sample $u\%$ of data from $\mathcal{D}_t$}
        \STATE{$\Phi = $ 30 randomly sampled configuration of hyperparameters of the $\mathcal{M}$}
        \FOR{each hyperparameter configuration $\phi_i \in \Phi$}
            \FOR{$j \in \{1,2,3\}$}
                \STATE{draw a random seed $r_j$}
                \STATE{$\mathcal{U}_t^j, \mathcal{U}_v^j =$ universe $\mathcal{U}$ split into training and validation subset using the random seed $r_j$}
                \STATE{$\mathcal{M}_{ij}$ = trained $\mathcal{M}$ using hyperparameter configuration $\phi_i$ on $\mathcal{U}_t^j$}
                \STATE{$a_{ij}$ = performance of $\mathcal{M}_{ij}$ evaluated on $\mathcal{U}_v^j$}
            \ENDFOR
        \STATE{$a_i = \frac{1}{3}\sum_{j=1}^{3} a_{ij}$}
        \ENDFOR
         \STATE{$\mathcal{A} = \{ a_i \hspace{0.5em} | \hspace{0.5em}  \forall \phi_i \in \Phi\}$}
        \STATE{$\mathcal{B} = \{ \mathcal{M}_{ij} \hspace{0.5em} | \hspace{0.5em} \forall a_i \in \text{top-3(}\mathcal{A}\text{)}   \}$}
        \STATE{$\mathcal{M}^* =$ an equally weighted ensemble of all models in $\mathcal{B}$}
        \STATE{$a^* =$ performance of $\mathcal{M}^*$ on $\mathcal{D}_s$}\\
        \STATE{\RETURN $\quad a^*$}
    \end{algorithmic}
\end{algorithm}

\newpage
\section*{Supplementary Note 6: ROI retrieval algorithm}
We describe the algorithm used to retrieve the ROI patches in Algorithm~\ref{alg:roi}. In all experiments, we set $H = W = 1024$, $h = w = 32$, $h_c = w_c = 256$, $\mathbb{T}_a = \{24, 48, 72, 96\}$, and $K = 6$.

\renewcommand{\algorithmicrequire}{\textbf{Input:}}
\renewcommand{\algorithmicensure}{\textbf{Output:}}
\begin{algorithm}
    \caption{ROI retrival}
    \label{alg:roi}
    \begin{algorithmic}[1]
        \REQUIRE  chest X-ray image $\mathbf{x} \in \mathbb{R}^{H,W}$, saliency maps $\mathbf{A} \in \mathbb{R}^{h,w,|\mathbb{T}_a|}$, number of ROI patches $K$
        \ENSURE a set of retrieved ROI patches $O = \{ \tilde{\mathbf{x}}_k |  \tilde{\mathbf{x}}_k \in \mathbb{R}^{h_c,w_c} \}$
        \STATE{$O = \emptyset$}
        \FOR{each time window $t \in \mathbb{T}_a$}
            \STATE{$\mathbf{\tilde{A}}^t = \text{min-max-normalization}(\mathbf{A}^t)$}
         \ENDFOR\\
         \STATE{$ \mathbf{A}^{*} = \sum_{t \in \mathbb{T}_a} \tilde{\mathbf{A}}^t$}
         \STATE{$l$ denotes an arbitrary $h_c \frac{h}{H} \times w_c \frac{w}{W}$ rectangular patch on $\mathbf{A}^{*}$}
         \STATE $\text{criterion}(l, \mathbf{A}^{*}) = \sum_{(i,j) \in l} \mathbf{A}^{*}[i,j]$
        \FOR{each $1,2,...,K$}
            \STATE{$l^* = \argmax_{l} \text{criterion}(l, \mathbf{A}^{*})$}
            \STATE{$L = $ position of $l^*$ in $\mathbf{x}$}
            \STATE{$O = O \cup \{L\}$}
            \STATE{ $\forall (i,j) \in l^*$, set $\mathbf{A}^*[i,j]=0$}
        \ENDFOR\\
        \STATE{\RETURN  $\quad O$}
    \end{algorithmic}
\end{algorithm}

\newpage
\section*{Supplementary Note 7: Learning curves}

\begin{figure}[H]
\renewcommand\figurename{Supplementary Figure}
    \centering
    
        \begin{tabular}{c c}
        \includegraphics[width=0.43\textwidth]{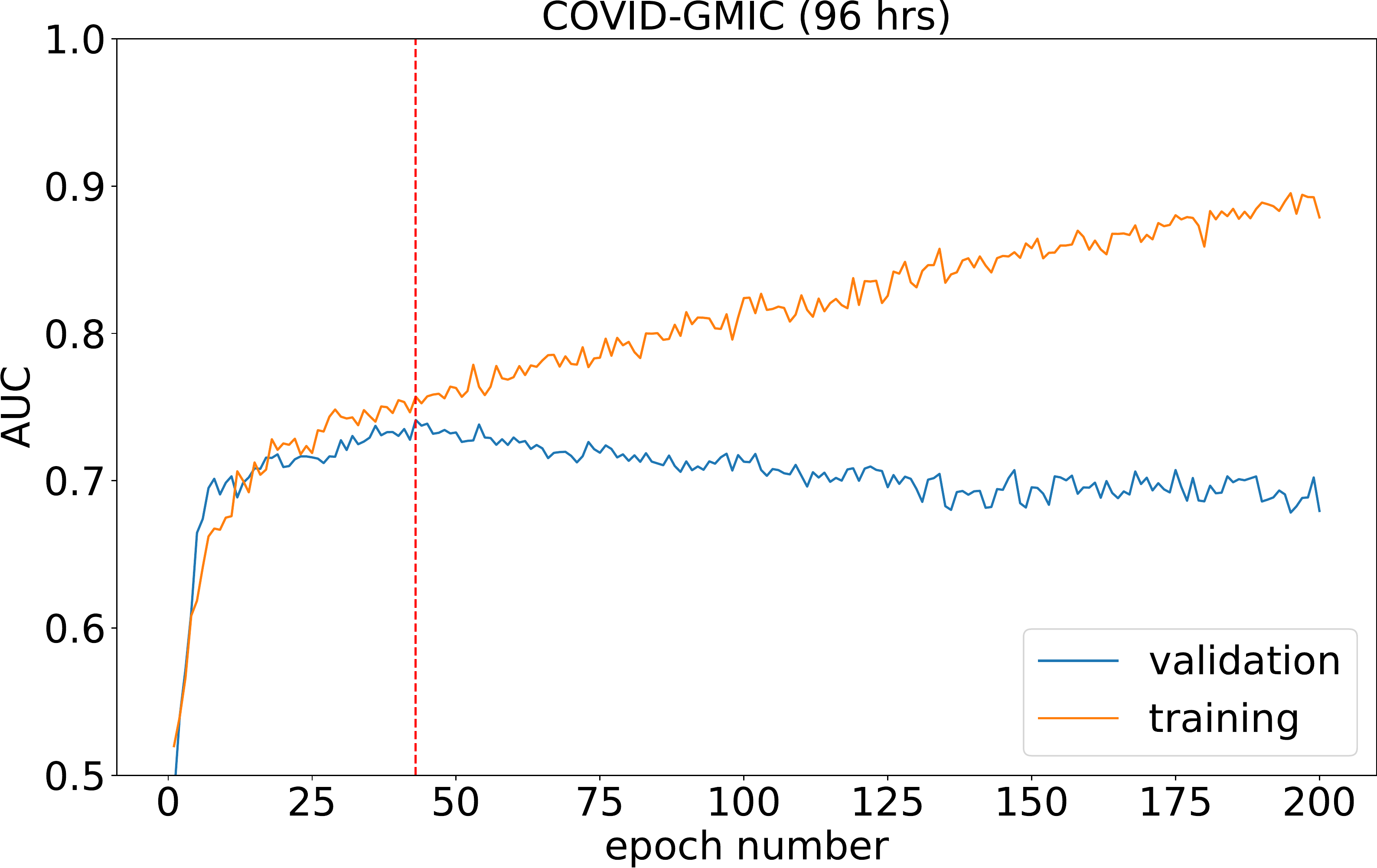}&
    \includegraphics[width=0.43\textwidth]{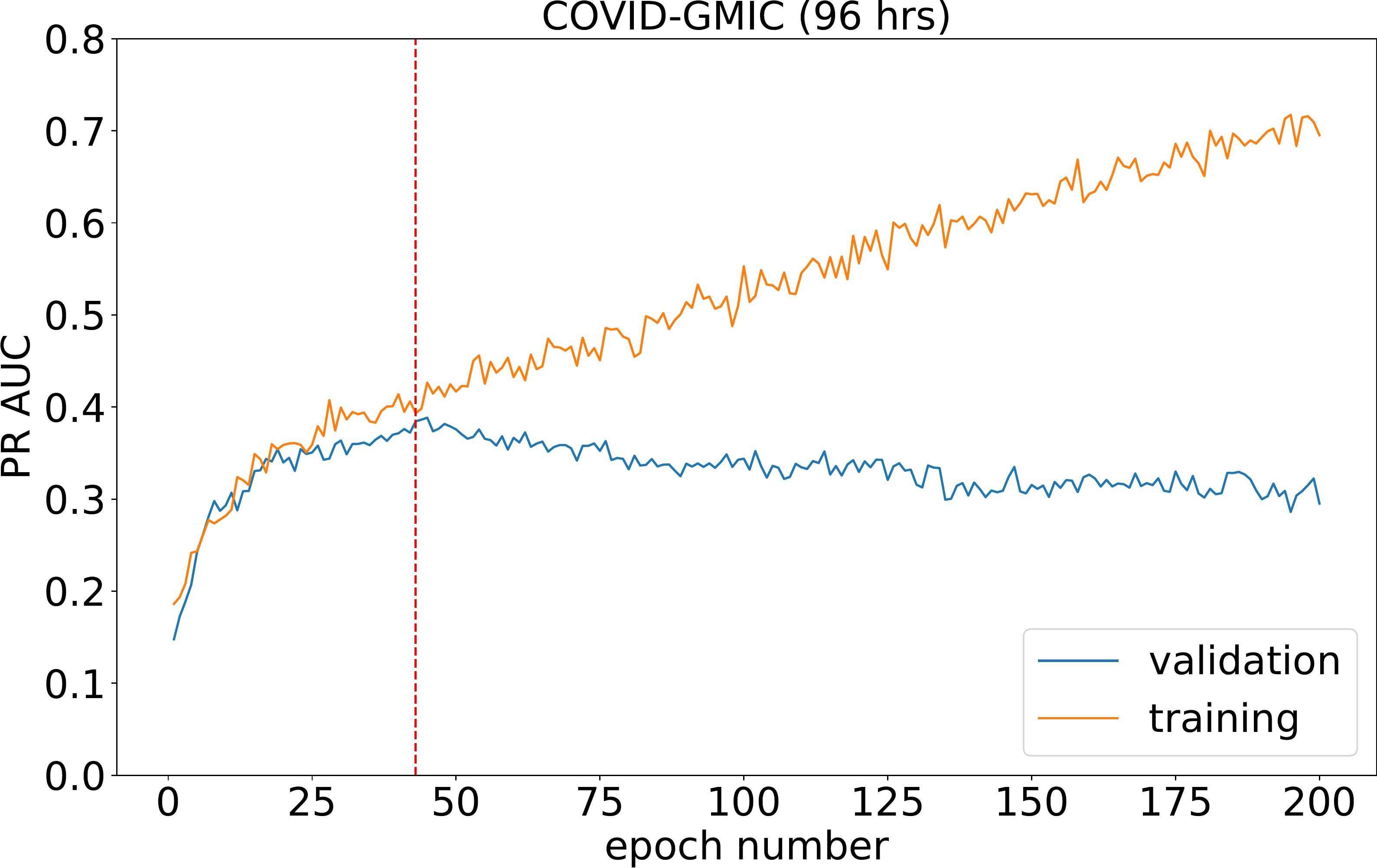}
         \\
        \includegraphics[width=0.43\textwidth]{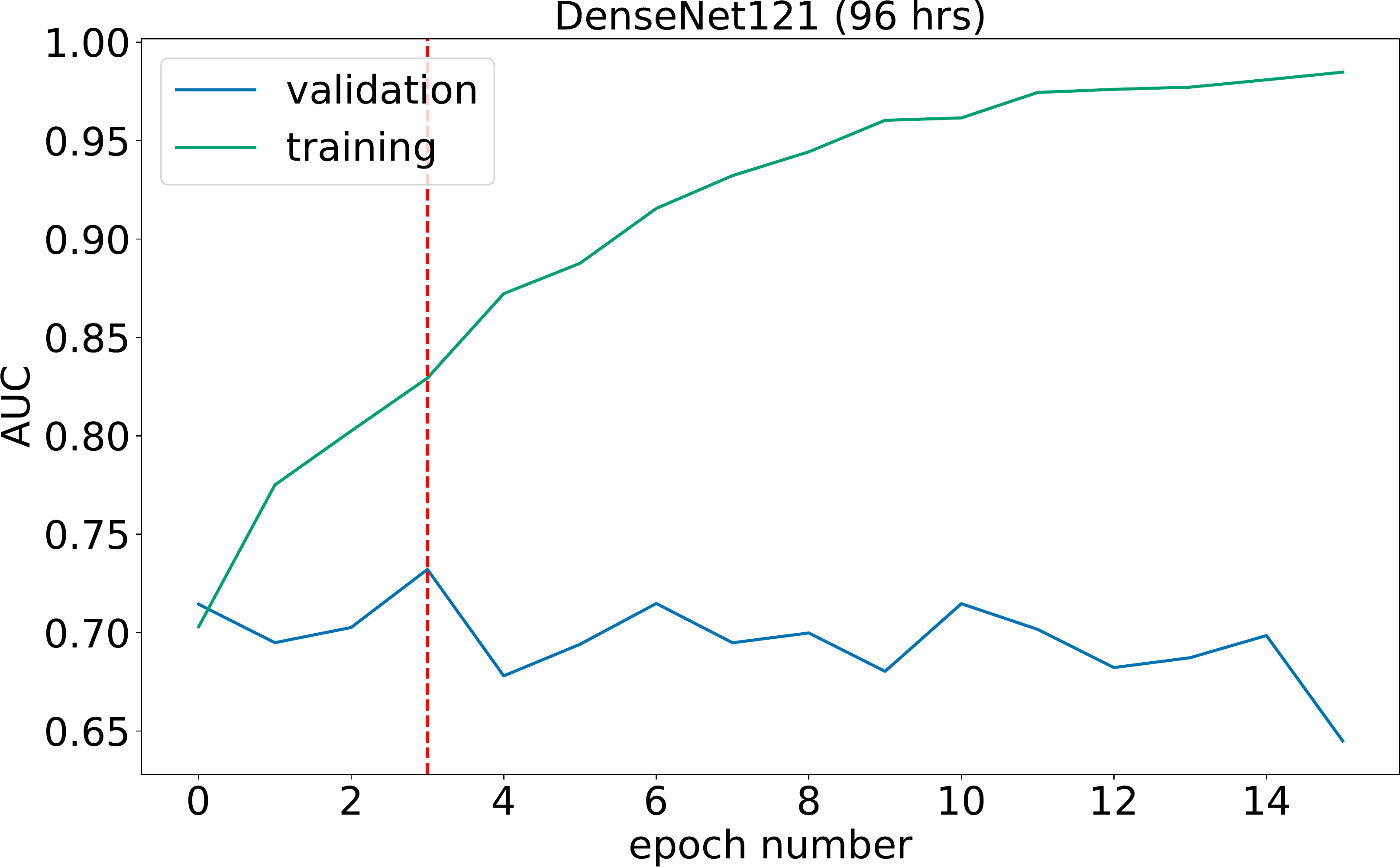} &
    \includegraphics[width=0.43\textwidth]{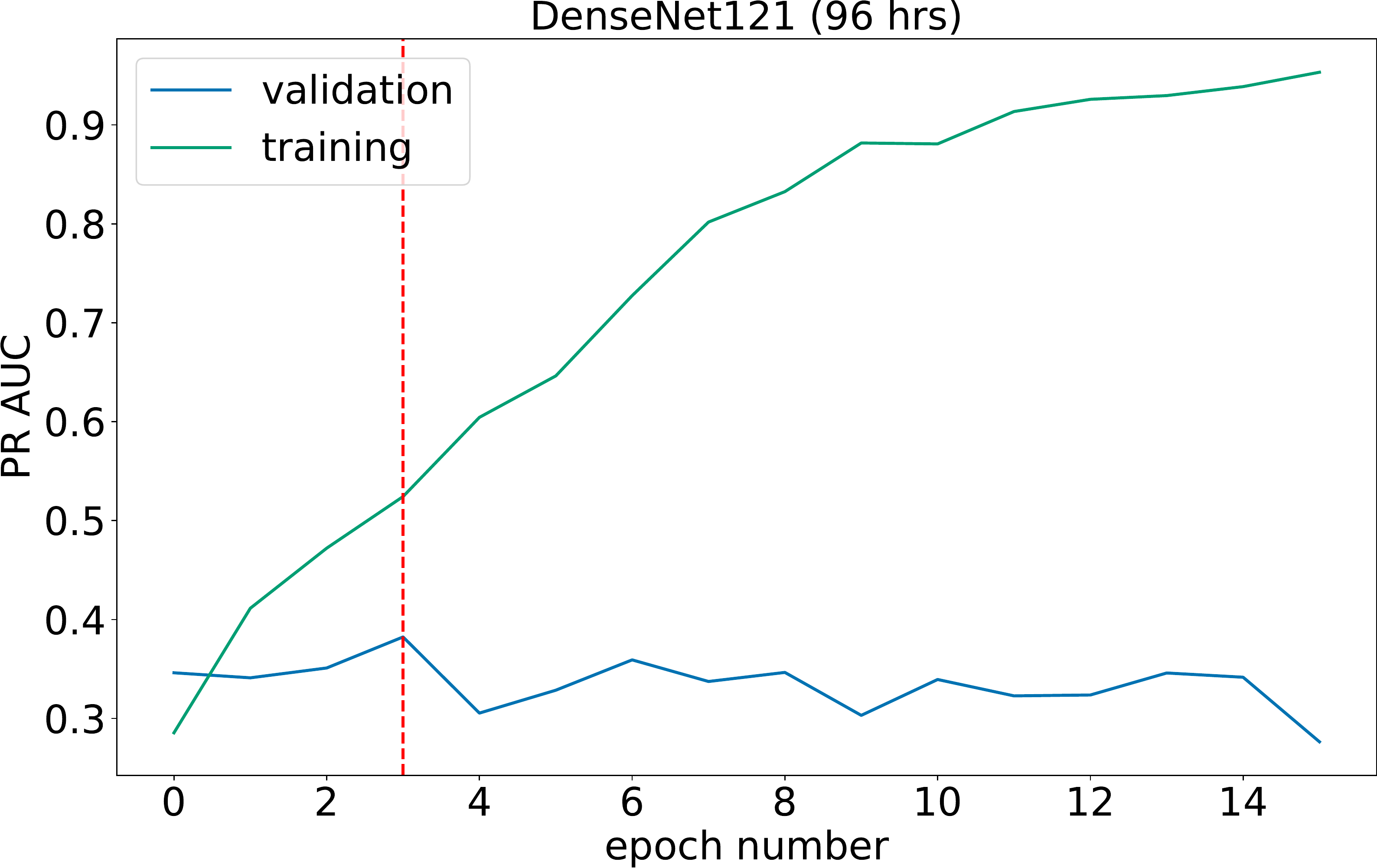} 
    \end{tabular}
    \caption{\small AUC and PR AUC for predicting clinical deterioration within 96 hours during training achieved by a selected COVID-GMIC model and a selected DenseNet121 model on the training and validation set. We select the best epoch (marked in red) in which the model achieves highest AUC on the validation set. Our model selection mechanism ensures that the selected model is sufficiently trained and does not lie in the overfitted regime. We observe similar trends in the learning curves for predicting clinical deterioration within 24, 48, and 72 hours.}
    \label{fig:learning_curves}
\end{figure}

\begin{figure}[H]
\renewcommand\figurename{Supplementary Figure}
    \centering
    \includegraphics[width=0.49\textwidth]{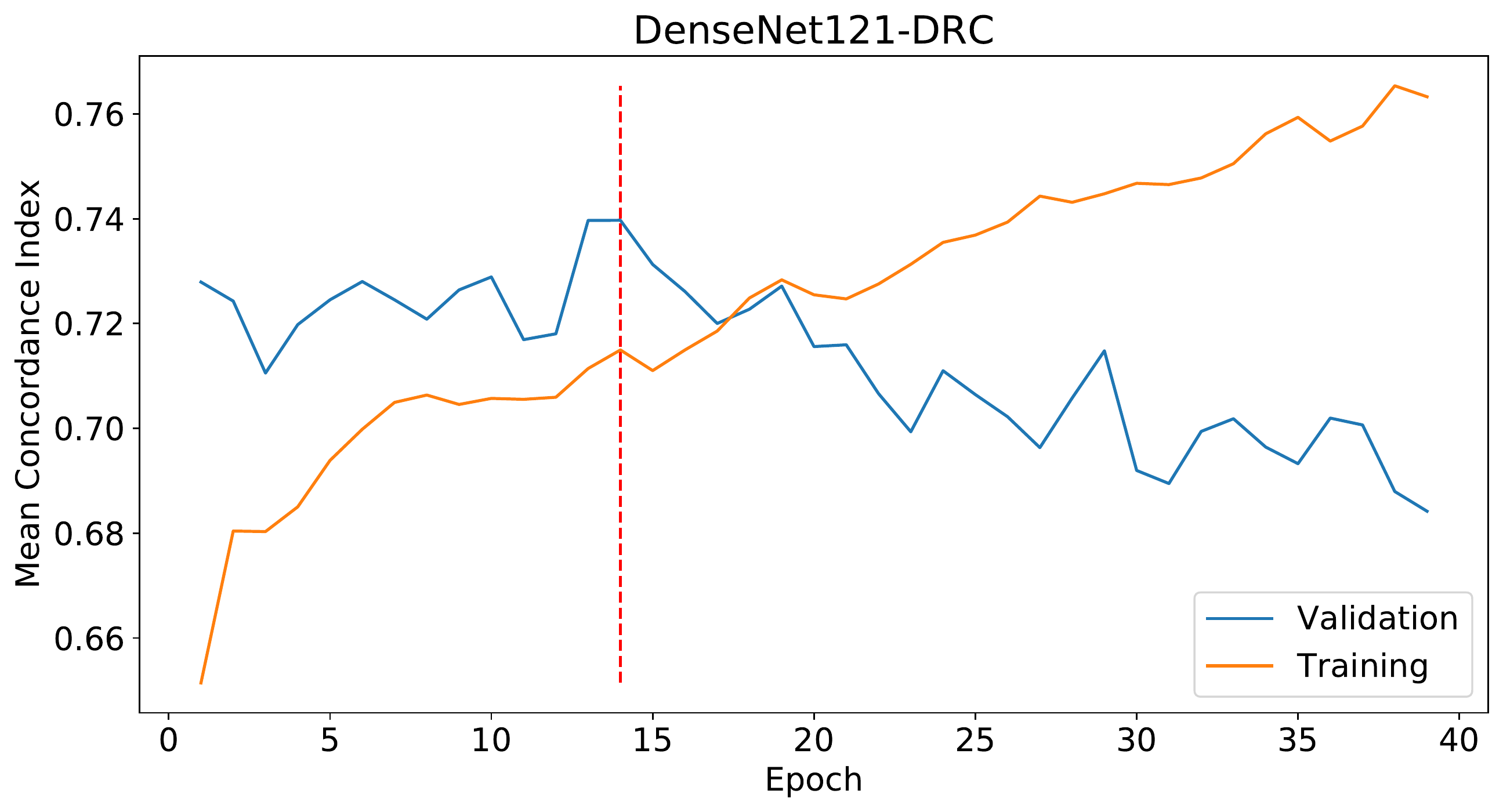}
    \includegraphics[width=0.49\textwidth]{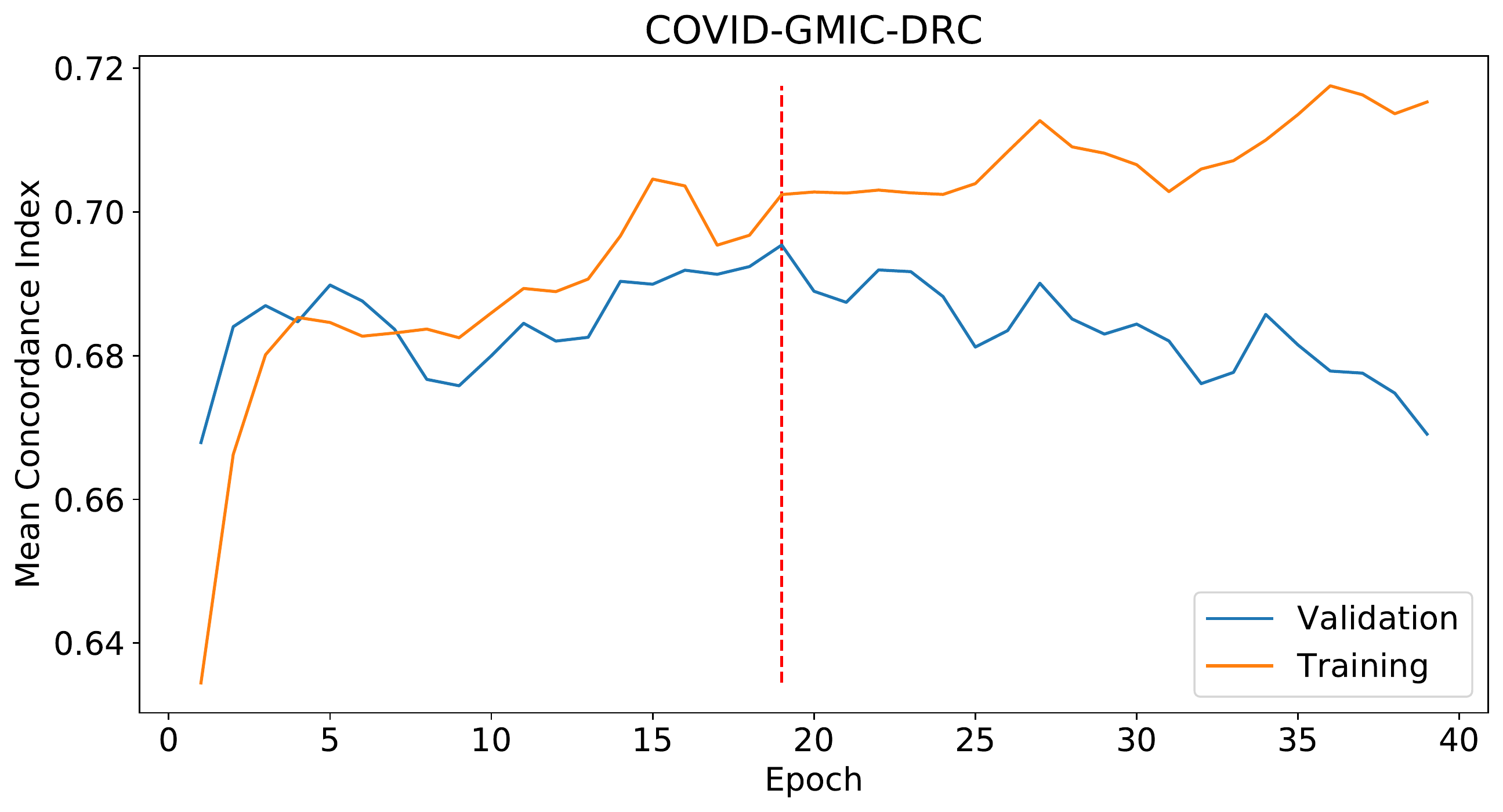}
    \caption{\small Training curves showing mean concordance index for DenseNet121 and COVID-GMIC-DRC models. We select the best epoch (marked in red) in which the model achieves highest mean concordance index on the validation set. Our model selection mechanism ensures that the selected model is sufficiently trained and does not lie in the overfitted regime.}
    \label{fig:learning_curves_drc}
\end{figure}

\end{document}